\RequirePackage{fix-cm}
\RequirePackage{rotating}
\documentclass[twocolumn]{svjour3}          
\smartqed  
\usepackage{graphicx}
\usepackage{amsmath}
\usepackage{amssymb}
\usepackage{booktabs}
\usepackage{epsfig}
\usepackage{bm}
\usepackage{makecell}
\usepackage{enumitem}
\usepackage{rotating}
\makeatletter
\@namedef{ver@everyshi.sty}{}
\makeatother
\usepackage{multirow}
\usepackage[normalem]{ulem}
\usepackage{ifsym}
\usepackage[pagebackref,breaklinks,colorlinks]{hyperref}
\usepackage[numbers]{natbib}

\usepackage{color}
\usepackage{contour}
\usepackage{textcomp}
\usepackage{adjustbox}
\usepackage[english,showlanguages]{babel}
\usepackage[protrusion=true, expansion=true, shrink=55, stretch=55, 
tracking=true, kerning=true, spacing=true, 
final]{microtype}

\contourlength{0.5pt}

\newcommand{\culine}[1]{%
  \uline{\phantom{#1}}%
  \llap{\contour{white}{#1}}%
}

\newcommand\blfootnote[1]{%
\begingroup
\renewcommand\thefootnote{}\footnote{#1}%
\addtocounter{footnote}{-1}%
\endgroup
}

\journalname{International Journal of Computer Vision}
\begin{document}

\title{3D Adversarial Augmentations for Robust Out-of-Domain Predictions}


\author{Alexander Lehner$^{*,1,2}$ \and
        Stefano Gasperini$^{*,1,2}$ \and
        Alvaro Marcos-Ramiro$^{2}$ \and
        Michael Schmidt$^{2}$ \and
        Nassir Navab$^{1}$ \and
        Benjamin Busam$^{1}$ \and
        Federico Tombari$^{1,3}$}

\authorrunning{Lehner and Gasperini, et al.} 

\institute{
$^*$ The authors contributed equally. \at \\
$^1$ CAMP, Technical University of Munich, Germany \\
$^2$ BMW Group, Munich, Germany \\
$^3$ Google, Zurich, Switzerland \\\\
Contact authors:\\
Alexander Lehner (\href{mailto:alexander.lehner@tum.de}{alexander.lehner@tum.de}) and\\
Stefano Gasperini (\href{mailto:stefano.gasperini@tum.de}{stefano.gasperini@tum.de}).
}


\maketitle

\begin{abstract}
Since real-world training datasets cannot properly sample the long tail of the underlying data distribution, corner cases and rare out-of-domain samples can severely hinder the performance of state-of-the-art models. This problem becomes even more severe for dense tasks, such as 3D semantic segmentation, where points of non-standard objects can be confidently associated to the wrong class. In this work, we focus on improving the generalization to out-of-domain data. We achieve this by augmenting the training set with adversarial examples. First, we learn a set of vectors that deform the objects in an adversarial fashion. To prevent the adversarial examples from being too far from the existing data distribution, we preserve their plausibility through a series of constraints, ensuring sensor-awareness and shapes smoothness. Then, we perform adversarial augmentation by applying the learned sample-independent vectors to the available objects when training a model. We conduct extensive experiments across a variety of scenarios on data from KITTI, Waymo, and CrashD for 3D object detection, and on data from SemanticKITTI, Waymo, and nuScenes for 3D semantic segmentation. Despite training on a standard single dataset, our approach substantially improves the robustness and generalization of both 3D object detection and 3D semantic segmentation methods to out-of-domain data.
\end{abstract}

\section{Introduction}

Reliable understanding of the surroundings in general settings is crucial for high automation~\cite{kilic2021lisa,lehner20223dvfield,mirza2022norm}.
However, current methods lack the necessary robustness and generalization capabilities to properly tackle unexpected events in safety-critical applications, such as autonomous driving and robotics~\cite{baier2019challenges}. Deploying state-of-the-art approaches directly to real-world problems raises a set of issues which go beyond solving the task on a public dataset.


\begin{figure*}[t]
\centering
  \includegraphics[width=1.0\textwidth]{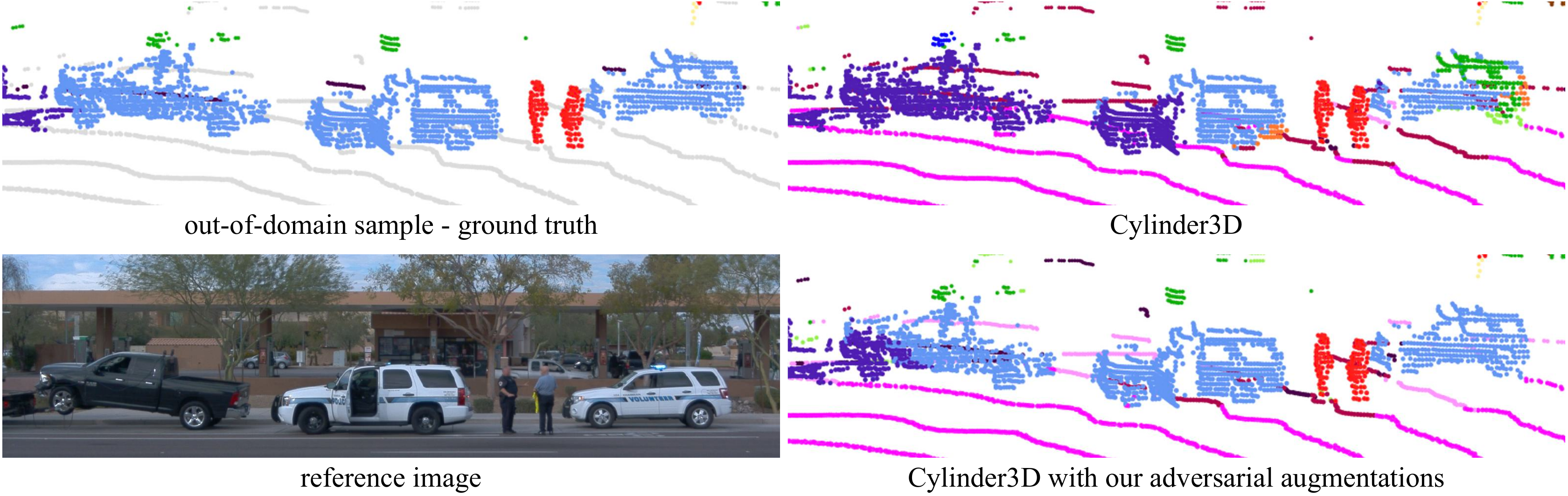}
   \caption{\protect\uline{3D semantic segmentation} predictions of Cylinder3D~\cite{zhu2021cylindrical} trained on SemanticKITTI~\cite{behley2019semantickitti} with and without our adversarial augmentations. This figure shows a challenging out-of-domain sample from Waymo~\cite{sun2020waymo}, including three large cars, of which a pickup truck being towed and a large SUV with an open door. Due to the domain gap, the standard Cylinder3D could not correctly segment any of the three vehicles in the scene. Instead, by plausibly expanding the available training data, our domain generalization method allowed the model to improve the segmentation of the out-of-domain cars. The road and other classes were ignored in the transfer to Waymo due to misaligned definitions across the datasets.}
   \label{fig:teaser_semantic}
\end{figure*}

In these setups, characteristics such as robustness and strong generalization become of utmost importance to circumvent dangerous consequences~\cite{gasperini2022holistic}.
As challenging scenes can drastically hinder performance~\cite{zhao2022oodcv}, particularly crucial is a model's ability to robustly generalize to unseen scenarios, e.g., out-of-domain and long tail samples, as well as to adverse illumination and weather conditions~\cite{gasperini_morbitzer2023md4all}, e.g., at night and with rain.
Various categories of works are aimed at mitigating these issues, including domain adaptation~\cite{yi2021complete_and_label}, domain generalization~\cite{wang2021generalizing}, simulations~\cite{beery2020synthetic}, estimating the uncertainty~\cite{gasperini2021certainnet}, and generating adversarial examples~\cite{tu_physically_2020}.

\begin{figure}[t]
\begin{center}
\includegraphics[width=1.00\linewidth]{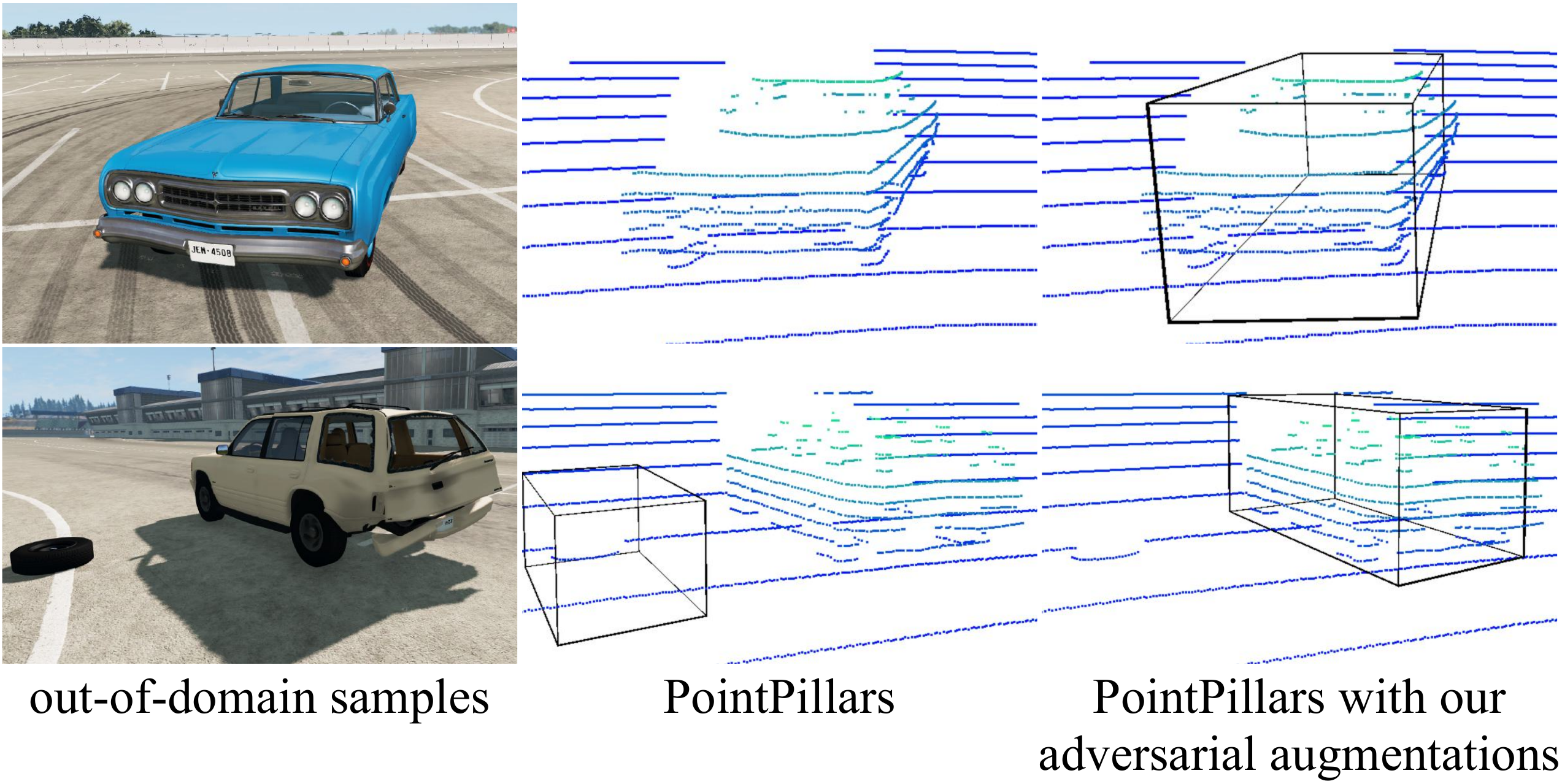}
\end{center}
   \caption{
   \protect\uline{3D object detection} predictions of PointPillars~\cite{lang_pointpillars_2019} trained on KITTI~\cite{geiger2012we}, without and with our adversarial augmentations, on out-of-domain cars from the proposed CrashD dataset. The models were transferred without any fine-tuning. Due to the unusual shapes of the vehicles compared to those in KITTI, the standard PointPillars could not detect them. By expanding the training distribution, our adversarial augmentations allowed the model to detect both the old car and the damaged one \hyperref[{sec:beamng}]{$^\Box$}.
   }
\label{fig:teaser_detection}
\end{figure}

Due to the difficulty of capturing corner cases and challenging situations from the long tail of the data distribution, training datasets cannot contain all possible scenarios~\cite{lehner20223dvfield} and typically include mostly average cases. This forces to design robust methods that work effectively not only on the available training data distribution, but also on rare unseen and unknown samples~\cite{gasperini2022holistic,jung2021sml}.
Since rare samples are typically unavailable at training time and might not be part of the same distribution, as shown in Figures~\ref{fig:teaser_semantic} and~\ref{fig:teaser_detection}, state-of-the-art approaches tend to fail with out-of-domain samples, thereby proving the need for more robust solutions~\cite{gasperini2022holistic,mirza2022norm}.

Although many works addressed part of these concerns on 2D image data~\cite{qiao2020learning, beery2020synthetic, gasperini2022holistic, hendrycks2021many, mirza2022norm, postels2019sampling}, these issues remain mostly open for 3D point clouds~\cite{lehner20223dvfield}.
Compared to using 2D data alone, 3D sensors (e.g., LiDAR and ToF cameras) offer an extra layer of redundancy and robustness, which is highly valuable in safety-critical settings.

In 3D semantic segmentation, methods assign a known class to each input point. Since 3D approaches heavily rely on the geometric relationship between 3D points, non-standard and out-of-domain objects, such as those in Figure~\ref{fig:teaser_semantic}, can be easily assigned to the wrong class, thereby posing a safety threat if not properly taken into account.
Moreover, since \textit{softmax} highly promotes the most probable class, current approaches tend to be extremely confident even with out-of-distribution samples, or when delivering wrong predictions~\cite{gawlikowski2021survey}, worsening the problem. As challenging scenarios can naturally occur in the real world~\cite{hendrycks2021natural}, robustness against them is a fundamental characteristic to ensure the safe deployment of a model.

To avoid the difficult and expensive collection of long tail samples for training, challenging scenarios can be designed artificially via adversarial attacks applied to readily available data~\cite{goodfellow_explaining_2015}. These adversarial examples expose the vulnerabilities of a model and can be incorporated in the training data to improve its robustness via adversarial augmentation~\cite{lehner20223dvfield, li2022adversarial_augm_pc_robustness}.
While existing methods have generated adversarial examples to improve robustness against out-of-domain data for 3D object detection~\cite{tu_physically_2020,lehner20223dvfield}, this remains still unexplored for 3D semantic segmentation.

In this work, we extend the capabilities of a model to out-of-domain data. We achieve this by enriching the training data with adversarial examples, which we make plausible via constraints for sensor-awareness and smooth deformations. First, we learn a small set of vectors, applicable to the entire dataset, to generate challenging samples. Then, we retrain the model while deforming available objects with our adversarial vectors, and integrate these adversarial examples as data augmentation. Thanks to difficult and plausible deformations, we substantially improve the generalization to challenging out-of-domain data, without using any extra information.
The main contributions of this work can be summarized as follows:
\begin{itemize}
    \item We raise awareness on natural adversarial examples, such as damaged and rare cars.
    \item We propose a sensor-aware and sample-independent adversarial augmentation method for domain generalization on 3D point clouds.
    \item We specialize our domain generalization approach for both 3D object detection and 3D semantic segmentation, making it the first effectively working on both tasks.
    \item We strengthen the model's decision boundaries via targeted and untargeted adversarial augmentation.
    \item We explain the impact of our method by analyzing the vectors learned for the two tasks.
    \item We publicly release CrashD: a dataset containing rare out-of-domain and long tail cars.
\end{itemize}

\protect\blfootnote{$^\Box$ \textit{Images used with courtesy of BeamNG GmbH.}\label{sec:beamng}}

Differently from the conference version of this work \cite{lehner20223dvfield}, here we: 1) extend our method to 3D semantic segmentation, which requires a different architecture and adversarial loss compared to 3D object detection (Section~\ref{sec:vec_field}); 2) generate our adversarial vectors both to resemble specific classes of choice (i.e., targeted attack) and also in an untargeted fashion; 3) assess the impact of the LiDAR intensity signal on the performance degradation of a model when transferring to different sensors (Section~\ref{sec:results_intensity});
4) generate adversarial intensity signals within our augmentations to mitigate the domain gap (Section~\ref{sec:method_intensity}); 5) generate adversarial augmentations of multiple object classes (Section~\ref{sec:anchor_points}); 6) eliminate the need for 3D bounding boxes exploiting point-level annotations; 7) analyze how our adversarial augmentations alter the decision boundaries of the model (Section~\ref{sec:results_main_generalization}); 8) assess the impact of our augmentations on the robustness against transformations of the input (Section~\ref{sec:results_intensity_robustness}); 9) analyze the differences between adversarial vectors learned for different tasks as well as their impact on the models (Section~\ref{sec:results_vectors_comp}); 10) train our vectors against a different architecture (Cylinder3D by~\cite{zhu2021cylindrical}); and 11) evaluate our method on three additional public datasets.


\section{Related Work}
\label{sec:related_work}
Our work focuses on adversarial augmentation for both 3D semantic segmentation and 3D object detection.
In this section, first, we provide an overview of 3D semantic segmentation methods (Section~\ref{sec:rw_semantic}), then we go through approaches addressing generalization across both tasks (Section~\ref{sec:rw_generalization}). Finally, we explore adversarial methods, focusing on 3D point clouds and semantic segmentation (Section~\ref{sec:rw_adversarial}).

\subsection{3D Semantic Segmentation}\label{sec:rw_semantic}
Depending on the input representation, semantic segmentation methods for LiDAR point clouds can be grouped in projection-based and point-based. The former project 3D point clouds to a regular grid, which allows to use standard convolutions. Regular grids can be voxels to use 3D convolutions~\cite{zhou2018voxelnet}, or pixels via spherical projections to use common and fast 2D convolutions well-known from the image domain~\cite{milioto2019rangenet++,wang2018pointseg}.
Voxels could be cubic~\cite{zhou2018voxelnet}, vertical square columns as in PointPillars~\cite{lang_pointpillars_2019}, or cylinder partitions as in Cylinder3D~\cite{zhu2021cylindrical}. Specifically, when partitioning the point cloud, Cylinder3D takes into account the increasing sparsity at higher distances, by means of distance-dependent voxel sizes.
Instead, point-based methods operate directly on the 3D point clouds, by extracting features exploiting the geometrical relationship between neighboring points.
PointNet was the first in this category~\cite{qi_pointnet_2017}, using multilayer perceptrons and aggregating the extracted features through a global max-pooling. Other works, such as KPConv~\cite{thomas2019kpconv}, focused on creating new operations dedicated to points.
More recently, a variety of hybrid approaches tried to exploit the benefits of both projection and point-based categories~\cite{tang2020searching,xu2021rpvnet}.

Other recent works explored the use of various aids, such as 2D data, attention mechanisms, sequential data and contrastive learning.
2DPASS~\cite{yan20222dpass} exploits 2D information to improve the representation learning on LiDAR point clouds.
(AF)$^2$-S3Net~\cite{cheng2021af2s3net} uses attention to capture fine details in LiDAR semantic segmentation.
\cite{chen2021movingobj} exploited sequential data to segment moving objects in LiDAR point clouds.
SegContrast~\cite{nunes2022segcontrast} improves representation learning in a self-supervised fashion via a contrastive loss, to drive similar structures of LiDAR point clouds towards each other in the embedding space.
LESS~\cite{liu2022less} aims at reducing human annotation for semantic segmentation of LiDAR point clouds, by leveraging geometrical patterns towards an heuristic pre-segmentation.
While 3D semantic segmentation assigns a semantic class to each point in input, 3D panoptic segmentation takes an extra step by additionally segmenting the points belonging to individual instances~\cite{gasperini2021panoster,marcuzzi2022contrastive_pano,razani2021gps3net_pano}.

In this work, we focus on 3D semantic segmentation and base our experiments on Cylinder3D~\cite{zhu2021cylindrical}, which we retrain with the only modification of including our generated adversarial examples at training time, to improve its generalization to out-of-domain data.

\subsection{Improving Generalization}\label{sec:rw_generalization}
Generalizing to unseen data is highly desirable for any learning-based model~\cite{wang2021generalizing}, particularly in unconstrained real-world settings, such as autonomous driving.
Unseen data comprises any sample that is not part of the training set, including data both in-domain (e.g., validation set) and out-of-domain (e.g., unseen categories).
Depending on the task and domain, a wide variety of techniques can be used to improve generalization and robustness, such as domain-robust sensor signals~\cite{gasperini2021r4dyn}, multi-modal fusion concepts~\cite{jung2021wild}, and robust 3D descriptors~\cite{tombari2010unique}.
Domain generalization aims at improving the performance on a target domain (e.g., data captured by a different sensor), without using any information about it~\cite{wang2021generalizing}. Instead, domain adaptation exploits knowledge about the target data~\cite{wang2020train, yi2021complete_and_label}.
Domain generalization methods can be categorized in two groups: those acting on the model itself, and those operating on the input data.

Among the former category, regularizing the model is commonly done to reduce overfitting~\cite{srivastava2014dropout} or address domain generalization~\cite{balaji2018metareg}. \cite{gasperini2021certainnet} found uncertainty estimation beneficial to reduce the domain gap between different data distributions. Since estimating the uncertainty provides information about the knowledge boundaries of a model, it is helpful to segment unknown objects belonging to unseen categories as well~\cite{gasperini2022holistic}. Furthermore, search algorithms were used to find new architectures that improve robustness~\cite{mok2021advrush}.

Generalization can be improved also by manipulating the input data. \cite{albuquerque2020improving} explored pretraining and multi-task learning to improve results on out-of-distribution data. Moreover, adding synthetic samples can strengthen the performance on rare classes~\cite{beery2020synthetic}.
Data augmentation~\cite{summers2019improved, zhang2021mixup, hendrycks2021many, nekrasov2021mix3d} is part of this category as well. Among these, adversarial approaches extend the training data with altered inputs learned in an adversarial fashion as a way to improve generalization~\cite{volpi2018generalizing, tu_physically_2020, qiao2020learning, lehner20223dvfield}.

The method we propose in this work addresses domain generalization (i.e., does not use any target information) and belongs to the data category, specifically to the adversarial approaches, detailed in Section~\ref{sec:rw_adversarial}.

\subsubsection{Generalization in 3D Object Detection}
In the context of generalization, some works addressed the task of 3D object detection.
In the image domain, \cite{simonelli2020towards} created virtual views normalizing the objects with respect to their distance. By doing so, they were able to generalize better to samples at different depths. \cite{tu_physically_2020} improved the generalization towards cars with roof-mounted objects, by using adversarial examples on LiDAR point clouds.
With LISA, \cite{kilic2021lisa} targeted adverse weather conditions with a physics-based simulator for LiDAR point clouds. They generated data and included it during training to improve the model robustness in challenging conditions.
\cite{wang2020train} explored domain adaptation to bridge the gap between LiDAR point clouds containing cars with different sizes, due to the distributions of vehicles in different countries (e.g., Germany and USA).

Generalization is also the focus of our work, where we explore adversarial augmentation for 3D object detectors on point clouds.

\subsubsection{Generalization in 3D Semantic Segmentation}
With the increasing interest of expanding the applicability of learning-based systems, several researchers aimed to improve generalization and robustness of 3D semantic segmentation approaches.
PCT~\cite{xiao2022transfer_syn2real_lidar} mitigates the reality gap from synthetic LiDAR data for semantic segmentation.
\cite{yi2021complete_and_label} explored domain adaptation for semantic segmentation of LiDAR point clouds. They used a network to recover the underlying point cloud surface, from which they transferred labels across different LiDAR sensors. They also executed an experiment without using target knowledge, thus performing domain generalization.
Mix3D~\cite{nekrasov2021mix3d} is a data augmentation technique for semantic segmentation aimed at generalizing beyond contextual priors of the training set. It works by combining two augmented scenes, thereby generating novel out-of-context environments.
\cite{li2022robust_lidar_seg} improved robustness by means of a test time augmentation strategy within a knowledge distillation framework, incorporating a Transformer-based voxel feature extractor.
The recent 3DLabelProp~\cite{sanchez2022domaingen} was the first method released for domain generalization in 3D semantic segmentation. Specifically, they assessed the generalization ability of state-of-the-art approaches and proposed a method exploiting both spatial and temporal information. By relying on the previous frames from a sequence, as well as the current input, they were able to increase the robustness of the output.

In this work, we improve the model robustness and generalization through adversarial augmentation. Unlike domain adaptation approaches which use knowledge about the target domain~\cite{yi2021complete_and_label}, similarly to 3DLabelProp we perform domain generalization, thus use solely information about the source domain. Moreover, our method operates on individual frames, instead of sequences as 3DLabelProp~\cite{sanchez2022domaingen}.
Therefore, our method differs from all existing approaches. Ours is also the first general approach shown to work effectively on both 3D object detection and 3D semantic segmentation.


\subsection{Adversarial Examples}
\label{sec:rw_adversarial}
Adversarial examples are input modifications purposely designed to lead a model to wrong predictions~\cite{szegedy_intriguing_2014, goodfellow_explaining_2015}. They are generated against a trained model (e.g., to reduce its accuracy) and are typically transferable to other models as well~\cite{lehner20223dvfield}.
A multitude of works already explored adversarial examples in the image domain~\cite{carlini_towards_2017, moosavi-dezfooli_deepfool_2016, papernot_limitations_2016, yuan_adversarial_2019, xiao_generating_2018}, where small pixel perturbations fool target models.
\cite{alaifari_adef_2019} deformed images via sample-specific adversarial vector fields (i.e., sets of vectors anchored to a set of points in a given space). \cite{wang_amora_2020} proposed adversarial morphing fields to alter image pixels spatially.
However, adversarial examples are still mostly unexplored on point clouds, especially those captured by 3D sensors (e.g., LiDAR and ToF camera).

\begin{figure}[t]
\begin{center}
\includegraphics[width=1.00\linewidth]{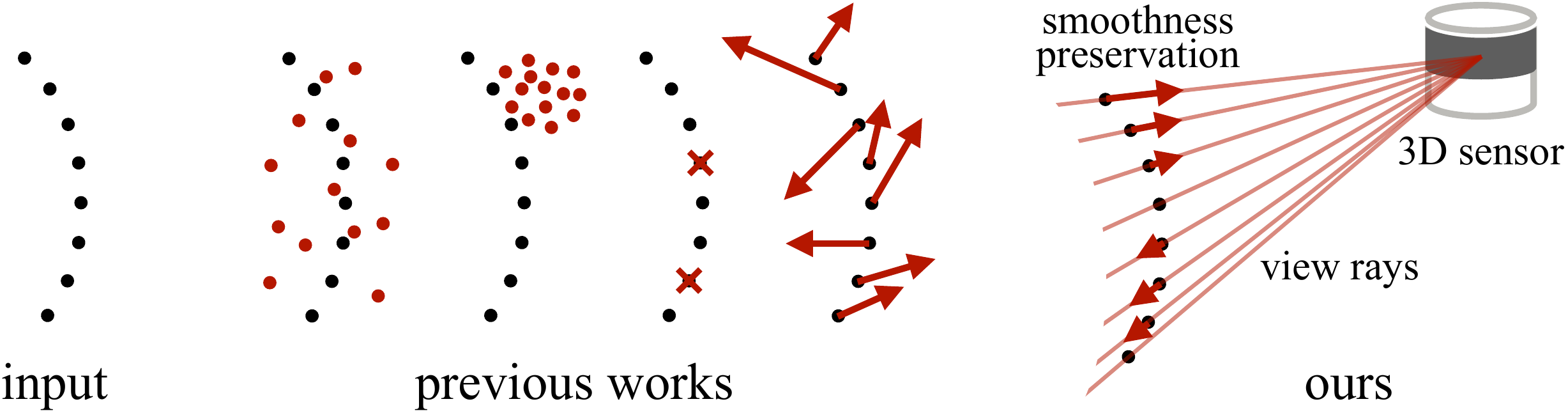}
\end{center}
   \caption{Adversarial perturbations introduced by previous works, compared to ours. While others add points, remove them, or shift them with minor constraints, our approach is sensor-aware and only slides points along the sensor view ray, thereby preserving occlusions. Additionally, our method produces adversarial examples with smooth surfaces, by considering the movement of the neighboring points when computing the shift for each point.}
\label{fig:constraints}
\end{figure}

\subsubsection{Adversarial Methods for Point Clouds}
Adversarial methods for 3D point clouds can be divided into three categories: generation if they add points, removal if they remove points, and perturbation if they only move points.

\textbf{Generation and removal}
\cite{xiang_generating_2019} were the first to work on adversarial point clouds and proposed a series of methods, some of which added points.
\cite{cao_adversarial_2019} added adversarial objects to LiDAR point clouds.
Similarly, \cite{tu_physically_2020} added adversarial meshes on the roof of cars.
A different set of works investigated sensor attacks, adding points with a spoofing device~\cite{cao_adversarial_2019-1}. Instead, removal methods learn to discard critical points in an adversarial fashion~\cite{yang2019adversarial_removal}.

\textbf{Perturbation}
\cite{xiang_generating_2019} proposed the first two perturbation methods. For the iterative gradient L2 attack, they adapted PGD from the image domain~\cite{madry2018pgd}, optimizing for a minimal deformation constrained by the L2 norm. They also proposed the Chamfer attack, using the Chamfer distance between the original and the deformed object to reduce the perceptibilty of the attack~\cite{liu2020adversarial}.
Our method is similar to the iterative gradient L2 attack, but we do not learn a dedicated vector for every point of each sample. Instead, we learn one sample-independent vector field (i.e., operating on the whole dataset) and introduce further constraints to improve the plausibility of our perturbations.
\cite{liu2020adversarial} explored more noticeable perturbations than the ones of \cite{xiang_generating_2019}. They produced continuous shapes by altering neighboring points accordingly.
\cite{cao_invisible_2021} generated adversarial objects and then 3D printed them to fool multi-modal (LiDAR and camera) object detectors.

\textbf{Generalization}
Most works on adversarial point clouds were proposed targeting the ModelNet dataset~\cite{xiang_generating_2019, liu2020adversarial, hamdi2020advpc}, which features a set of synthetic 3D point clouds with various object shapes. As the samples of ModelNet were not captured by a 3D sensor, these pioneering works often produce unrealistic outputs~\cite{xiang_generating_2019, liu2020adversarial}. In fact, these approaches were not intended to improve the generalization of the models, but rather set the basis for adversarial attacks on point clouds~\cite{xiang_generating_2019}. Additionally, they are all sample-specific, making their applicability limited in practice, as they would require to be optimized independently for each object they are applied on~\cite{xiang_generating_2019, liu2020adversarial, hamdi2020advpc}.
Instead, \cite{tu_physically_2020} assessed the impact on 3D object detection of meshed objects (e.g., canoes and couches) synthesized on top of a car roof within a LiDAR scene. Moreover, they attacked these meshes and used them to defend the detector. By doing so, they improve robustness and generalization to unseen cars with roof-mounted objects.

Our work is significantly different from all sample-specific methods~\cite{alaifari_adef_2019, xiang_generating_2019, liu2020adversarial, yang2019adversarial_removal}. In fact, we construct a single set of vectors.
Similarly to \cite{tu_physically_2020}, we aim to improve the generalization to out-of-domain data. However, unlike them, as can be seen in Figure~\ref{fig:constraints}, we do not add any points, making ours a perturbation method.
Additionally, by not making any assumptions on the object nor the kind of sensor, our method has a wider applicability than that of \cite{tu_physically_2020}. Moreover, we preserve the plausibility of our adversarial examples by considering occlusion constraints, which were ignored so far, and making our perturbations sensor-aware. We achieve this by shifting points only along the sensor ray. Furthermore, our method differs from previous works as it generates adversarial examples via transferable learned vector fields.

In the conference version of this work (\cite{lehner20223dvfield}) we first deformed objects to fool a 3D object detector, then incorporated these adversarial examples when training a new detector (adversarial augmentation). By doing so, we substantially improved its robustness and generalization to out-of-domain samples.
In this work, we aim at extending this to 3D semantic segmentation. Moreover, we bring a series of improvements towards out-of-domain generalization, such as targeted adversarial augmentation, as well as new insights on our method and domain generalization via thorough analysis.

\begin{figure*}[t]
\begin{center}
\includegraphics[width=1.00\textwidth]{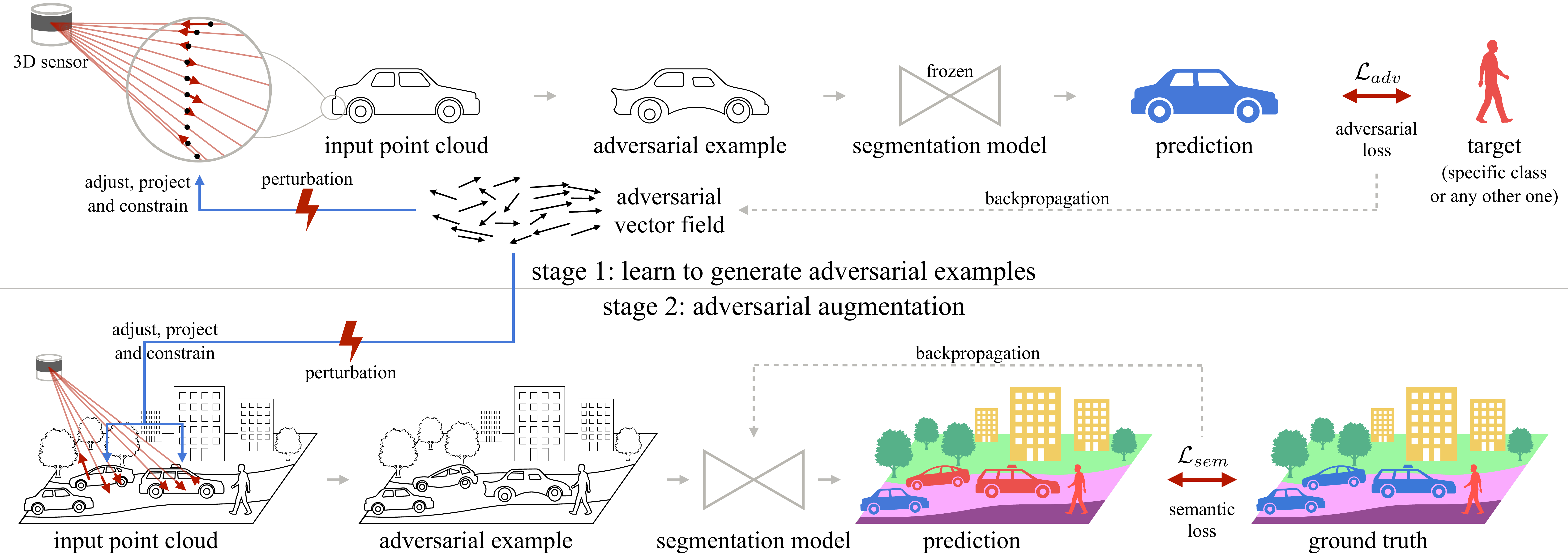}
\end{center}
   \caption{Overview of the proposed method. In the first stage (top), we \protect\culine{learn a vector field} by iteratively deforming objects to minimize the adversarial loss $\mathcal{L}_{adv}$ against a frozen model. The applied deformations are plausible thanks to a series of constraints (e.g., moving points only along their sensor view ray), and they can either target a specific different class (e.g., \textit{person} as in the figure), or remain untargeted to any other class.
   In the second stage (bottom), the vector field learned in the first stage is deployed throughout the scenes for \protect\culine{adversarial augmentation}. The augmented scenes are used during the training of a new robust segmentation model, thereby improving generalization to unseen objects. While the diagram exemplifies the process for 3D semantic segmentation, the same method can be similarly applied to other tasks, such as 3D object detection.}
\label{fig:framework}
\end{figure*}

\subsubsection{Adversarial Methods for Semantic Segmentation}
The majority of work exploiting adversarial examples (e.g., as adversarial attacks or for data augmentation) has focused either on the imaging domain~\cite{luo2021adversarial_semantic_img,abdollahi2021GAN_sem_img,arnab2018robustness} or on different tasks, such as classification~\cite{li2022adversarial_augm_pc_robustness} or object detection~\cite{tu_physically_2020,lehner20223dvfield}. Instead, only few methods were proposed for semantic segmentation of point clouds.
Among these, AttAN~\cite{zhang2021attan} explores adversarial learning to improve the predictions by correcting noisy ones, while \cite{zhu2021adversarial_lidar_semantic} proposed an adversarial attack in which they carefully added real objects in a LiDAR scene, to fool state-of-the-art segmentation methods.

Our work generates adversarial examples and performs adversarial augmentation for 3D semantic segmentation, substantially improving the generalization to out-of-domain data by expanding the available training set with hard and plausible examples.


\section{Method}
Our method is based on deforming point clouds to improve generalization and robustness against natural object variations and out-of-domain data. We achieve this via adversarial augmentation. We apply our approach both on 3D object detection and 3D semantic segmentation.
As shown in Figure~\ref{fig:framework}, our method is based on a vector field learned in an adversarial fashion.
After training the vectors against a frozen target model (Section~\ref{sec:vec_field}), we freeze the vector field and apply it to objects in the available training data. The same vector field can be applied to any seen or unseen object. To do so, we first scale the vector field to match the target object dimensions. Then, we constrain the points movement to preserve the object shape and occlusions, while making the deformations sensor-aware (Section~\ref{sec:field_application}). Our vector fields are class specific and we use them to deform all objects of their class. Such deformed objects are adversarial examples, which we integrate during training as data augmentation (Section~\ref{sec:augmentation}).

\subsection{Adversarial Vector Field}
\label{sec:vec_field}

Since the goal is fooling the detector/classifier by perturbing a point cloud without adding nor removing points, vectors are convenient in this setting, because they represent shifts in the point cloud directly. Additionally, using vectors allows for both compactness and transferability, as the same learned vector field can be applied to any target object.

\textbf{Construction} To create a lattice of uniformly spaced 3D vectors within a 3D bounding box, we discretize the space of a default bounding box $B_o$ with a step size $t$. This generates root coordinates $\bm{f}$ in 3D space, each of which is assigned an empty vector $\bm{v}= (x,y,z)$. The default bounding box $B_o$ is defined by its width $w$, height $h$, length $l$, orientation angle $\alpha$, and its center $\varsigma=(x,y,z)$.

\subsubsection{Adversarial Losses}

\textbf{For 3D object detection}\label{sec:adv_loss}
To suppress irrelevant bounding box proposals and focus on the ones that are most likely to be accurate, we use a binary cross entropy loss following the method described by~\cite{tu_physically_2020}.
Let $\mathcal{Q}$ be the set of relevant proposals, where each proposal $q$ has a confidence score $s$.
A proposal is considered relevant if its prediction confidence score $s > 0.1$.
We minimize $s$ by weighting it according to the 3D IoU with the ground truth $q^*$. The objective is defined as:
\begin{align}
\mathcal{L}_{adv.od} = \sum_{\substack{q, s~\in~\mathcal{Q}}} - \mathrm{IoU}(q^*, q) \log(1 - s).
\end{align}
During training, minimizing $\mathcal{L}_{adv.od}$ optimizes the vector fields to reduce the confidence score of the prediction of the detector. As the optimization converges, this causes the detector to either miss the object or predict a misaligned bounding box.

\textbf{For 3D semantic segmentation} Specifying our method to the task of 3D semantic segmentation, we explore two alternative configurations, namely untargeted and targeted adversarial augmentations. As for 3D object detection, we first learn a set of vector fields, via a loss function which depends on the configuration, as described below.
While augmenting with untargeted adversarial examples aims at strengthening the weakest decision boundaries of a model, using targeted ones gives the option of reinforcing a specific class boundary of interest.

\textbf{Untargeted loss for 3D semantic segmentation}
The untargeted attack aims to change the model's prediction $\rho$ of each point to any wrong class. This attack is trained by maximising the cross-entropy loss which would be normally minimized when training the segmentation model. Let $C$ be the number of classes, $\gamma_{\bm{p},c}$ be the binary indicator (0 or 1) if class label $c$ is the correct classification for its points $\bm{p}$ and $\rho_{\bm{p},c}$ the predicted probability of all points $\bm{p}$ belonging to class $c$. The objective function to be minimized for the untargeted vectors is defined as the opposite of the standard cross-entropy loss as:
\begin{align}
\mathcal{L}_{adv.ssu} = \sum^{C}_{c=1} \gamma_{\bm{p},c} \log(\rho_{\bm{p},c}).
\end{align}
So the goal is to fool the model into wrongly classifying as many points as possible. The easiest way to minimize $\mathcal{L}_{adv.ssu}$ is attacking the weakest decision boundary for each point. This loss formulation aims at globally degrading the performance of the model, also beyond the points being perturbed directly (i.e., belonging to the class of interest). Therefore, it can shift points to change the predictions of other points of which it does not change the location.

\textbf{Targeted loss for 3D semantic segmentation}
The targeted attack aims at changing the model's predictions to a specific class. So instead of maximizing the loss of the correct class as in the untargeted setting, here we minimize the loss for a different specific class of choice $\hat{c}$. Let $\bm{p}_{{c}^*}$ be the points of the class $c^*$ whose points are perturbed (i.e., adversarial class), and $\rho_{\bm{p},{c}^*}$ the model's predictions for these points. Given $\gamma_{\hat{c}}$ as the binary indicator for the chosen target class $\hat{c}$, the objective function to be minimized for our targeted vectors is defined as:
\begin{align}
\mathcal{L}_{adv.sst} = - (\gamma_{\hat{c}} \log(\rho_{{c}^*})).
\end{align}
This formulation forces the adversarial vectors to cross the decision boundary between the correct class ${c}^*$ and the target class $\hat{c}$. Unlike the untargeted loss $\mathcal{L}_{adv.ssu}$ which aimed at reducing the IoU across all classes, by acting upon the points of the adversarial class, the targeted loss $\mathcal{L}_{adv.sst}$ ignores the predictions on all points other than those of the adversarial class ${c}^*$.

\textbf{Training procedure} To train the vector field, we apply the same vectors to all target objects in every scene in the training set, and minimize the adversarial loss across the entire dataset. This process iteratively updates the vectors, resulting in different deformations of the target objects and ultimately leading to different predictions. As the chosen adversarial loss $\mathcal{L}_{adv}$ smoothly converges, the performance of the model, against which the vector field is optimized, decreases. Once the vectors have been trained, they can be used for data augmentation to improve the performance of a new model.

\subsubsection{LiDAR intensity}\label{sec:method_intensity} For models using the LiDAR intensity (also called reflectivity, or reflectance), such as Cylinder3D~\cite{zhu2021cylindrical}, we perturb not only the 3D location of the LiDAR points, but also their intensity signal $\tau$. In these cases, we add an additional dimension to our adversarial vectors, making them 4-dimensional, i.e., 3 spatial coordinates plus intensity, as $\bm{v}= (x,y,z,\tau)$. By doing so, the adversarial loss affects not only the 3D location of the points, but also their intensity values. Therefore, the same loss functions $\mathcal{L}_{adv}$ can be applied to learn these 4D vectors.

\subsection{Point Cloud Perturbation}
\label{sec:field_application}
To apply a vector field, we first scale it to fit the target object's size. We then manipulate the points using these vectors and constrain their movement as described in the following sections.

\subsubsection{Anchor points}\label{sec:anchor_points}
By being sample-independent, our vectors must maintain a relative spacing between them. Therefore, each vector needs to be anchored to a certain point, which can be either part of the point cloud, or not. This is crucial both when learning and when applying the vectors, otherwise it would be unclear how to move which points.
In the conference version of this work, where we focused only on 3D object detecion (\cite{lehner20223dvfield}), we anchored the vectors to a grid formed within a reference bounding box $B_o$. Specifically, we relied on ground truth 3D bounding boxes to learn and apply our vector fields as augmentation. However, for 3D semantic segmentation, 3D bounding boxes may not be available.
A naive approach without utilizing 3D boxes could employ a large vector field that covers the entire scene, which could be active only on the points belonging to a certain adversarial class (e.g., \textit{car}). This has severe drawbacks: it would require a very high amount of vectors, and the vectors could overfit due to the limited samples available at certain locations within the scene (e.g., at far distances), reducing their efficacy.
Instead, we address the lack of ground truth 3D bounding boxes with two alternative solutions: an off-the-shelf 3D object detector to predict 3D bounding boxes, and axis-aligned bounding boxes around the points belonging to each instance, which are even simpler to obtain.

\textbf{Predicted bounding boxes}
Using 3D bounding boxes when applying our method on either 3D object detection or 3D semantic segmentation has the benefit of following a similar pipeline for both tasks. While ground truth bounding boxes are readily available on datasets designed for 3D object detection (e.g., KITTI~\cite{geiger2012we}), they may be unavailable for semantic data (e.g., SemanticKITTI~\cite{behley2019semantickitti}). In the latter case, they can be obtained with an off-the-shelf detector (e.g.,~\cite{deng2021voxelrcnn}). However, the effectiveness of our method would then depend on the performance of the 3D object detector. To mitigate this dependency, we discard false positives by ensuring that all boxes contain points annotated with the correct class. When learning to deform, false negatives can be problematic because if too few samples are detected, the vectors may overfit. When performing adversarial augmentation, too many false negatives render the augmentation useless as it would be applied only on a limited number of samples, which would not allow for improved generalization to out-of-domain data.
We use predicted boxes for the popular \textit{car} class. Instead, for smaller classes, such as \textit{person}, the performance of 3D object detectors is not as reliable (also due to the KITTI dataset~\cite{geiger2012we} missing annotations for \textit{person} and \textit{cyclist}), requiring different solutions.

\textbf{Axis-aligned boxes}\label{sec:axis_aligned}
Using axis-aligned boxes instead of predicted boxes implies having access to point-level instance annotations. We construct an axis-aligned box around the points belonging to each instance. Compared to predicted boxes, using ground truth instance annotations guarantees the correct coverage of all objects. The drawback lies in having larger grids (e.g., when the object is rotated by 45° with respect to the ground axes), which causes a sparser distribution of vectors on the points to be perturbed. However, this strategy has the advantage that when applying the vector fields all objects can be deformed, assuming that their points are annotated.
Since existing 3D object detectors for LiDAR data mainly focus on the \textit{car} class, which is the one having the most accurate annotations in KITTI~\cite{geiger2012we}, their performance on other classes is sub-optimal. Therefore, for classes other than \textit{car} (e.g., \textit{person}) we opt for axis-aligned boxes. Since these boxes have no direction indication, we extract a pseudo orientation, by considering the shorter side $min=(w,l)$. Such ambiguity prevents the vector fields to specialize to certain orientations. Nevertheless, we deal with the direction ambiguities as described in Section~\ref{sec:rel_rot}.

\begin{figure*}[t]
\centering
  \includegraphics[width=1.0\linewidth]{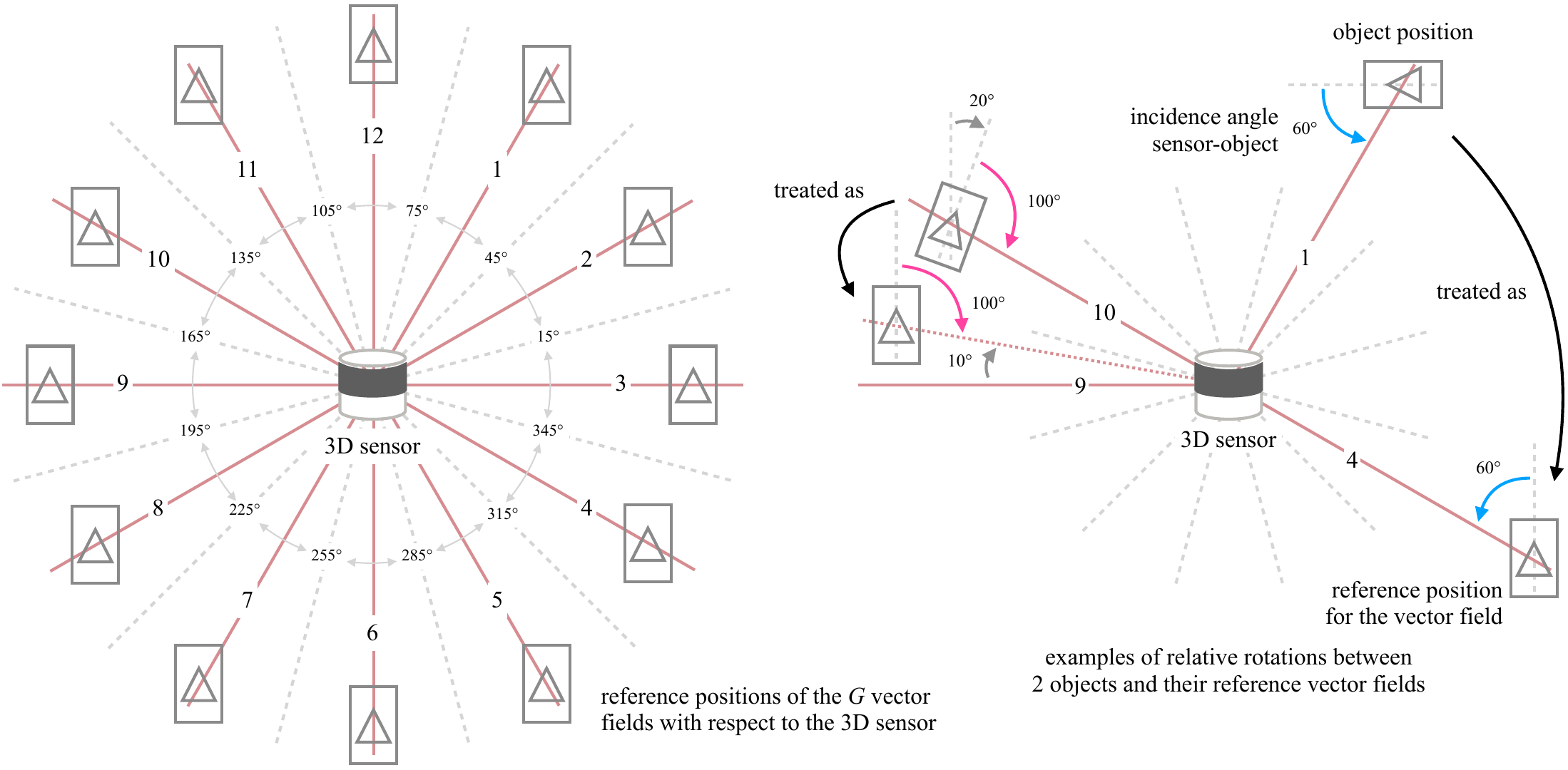}
   \caption{Representation of the relative rotations between the 3D sensor and the objects. The triangle inside each box points towards the direction of the object (e.g., a \textit{car} driving away from the sensor at position 12, left). We use $G=12$ vector fields, one for each of the 12 positions in the scheme (left). The red sensor rays are positioned at the $\beta_g$ angles, such as 0, 30, and 60 degrees. As we consider relative rotations, objects pointing in other directions are treated as axis-aligned objects at different positions (left) depending on their incidence angles from the sensor rays. For example (right), the object at position 1 pointing left is hit by the sensor ray with an incidence angle of 60° (blue), so it will be treated by the 4-th vector field, which will keep its incidence angle unchanged whilst rotating the object by 90° clockwise such that it points forward. Moreover, objects at angles that do not match the $G$ reference positions are considered at intermediate positions according to their rotation angles. So the object at position 10 rotated clockwise by 20° is treated by the 9-th vector field at an angle that is 10° clockwise from the reference position 9, thereby keeping the incidence angle unchanged at 100° (pink).}
   \label{fig:relative_rotation}
\end{figure*}

\subsubsection{Plausibility Constraints}\label{sec:constraints}
\textbf{Optical ray consistency}
To improve the generalization ability of our models and to ensure that the deformations take into account the physical constraints of the sensor, we use a sensor model that allows 3D points to be moved only along the optical ray. To do this, we first compute the ray $\bm{u}_i$ between the sensor and each point $\bm{p}_i$, which determines the direction in which each point can be moved. Then, we calculate the deformation vectors $\bm{r}_{i}$ for each point $\bm{p}_i$ by projecting its nearest vector $\bm{v}_{i}$ onto the ray $\bm{u}_i$. This means that each point can only be moved by the vector $\bm{r}_{i}$.

\textbf{Regularizing the deformations}
To limit the amount by which the points can be perturbed, we restrict the vectors $\bm{v}_{i}$ to have a maximum magnitude of $\epsilon$ according to the standard PGD $L_\infty$ attack~\cite{madry2018pgd}. We also ensure that the resulting deformed shape has smooth surfaces by sampling multiple neighboring vectors and using them to move a given 3D point. For each $j$-th vector among the $k$ nearest neighbors, we calculate the Euclidean distance $d_{ij}$ between the point $\bm{p}_i$ and its nearest vector $\bm{v}_{ij}$ from the vector field.
The final position of each point is determined by weighing the deformation vectors $\bm{r}_{ij}$ with their corresponding distances $d_{ij}$ and summing them together:
\begin{align}\label{eq:neighbor}
\bm{m}_i = \frac{\sum_{j=1}^{k} \frac{1}{d_{ij}} \bm{r}_{ij} }{\sum_{j=1}^{k} \frac{1}{d_{ij}}}
\end{align}
This allows for a smoother transition in depth between neighboring points, as vectors with opposite directions would cancel each other out and result in little or no movement of the affected point. All together, this helps to preserve the smoothness of the deformed shape and reduce the amount of irregular deformations.

\textbf{Regularizing the intensity shift}
When using the intensity as extra input signal, the forth dimension of each vector is used to increase or decrease its value $\tau$. Since the intensity is a scalar, it is independent from the optical ray direction. However, we constrain the adversarial shift $\Lambda$ on the intensity values $\tau$ for each point $\bm{p}_i$ by a maximum of $\psi$. To ensure smooth intensity changes on neighboring points we also weight the final intensity perturbation similarly to Equation~\ref{eq:neighbor}. The resulting intensity value is then clipped within the range $[0,1]$.

The effect of our attack is illustrated in Figure~\ref{fig:qualitative_attack_semantic}, where it can be seen how compared to the sample-specific Chamfer attack~\cite{liu2020adversarial}, ours preserves the overall object shape, moving the points only slightly.

\subsubsection{Relative rotation}\label{sec:rel_rot}
While it is possible to use a single vector field for all objects of a certain class in the entire dataset, this would lead to rather small and generic deformations. By fitting the same vector field to all objects, its vectors need to point in all directions to be able to shift points across the highly diverse object poses with respect to the capturing sensor. This wide variability of directions reduces its efficacy, resulting in small deformations.
We circumvent this problem and allow for a larger degree of alignment between neighboring vectors, by learning $G$ different vector fields according to the relative rotation of the object to the sensor along yaw.
Specifically, as shown in Figure~\ref{fig:relative_rotation} from a top-down view, we divide the surrounding 360° angle around the vertical axis centered at the 3D sensor in $G=12$ slices of 30° each, starting from 15°. By doing so, each vector field $g$ focuses on perturbing only objects visible from $\pm$ 15° around $\beta_g$, which is the reference angle for the vector field $g$ (shown in Figure~\ref{fig:relative_rotation} by the red sensor rays). In particular, as illustrated in the figure, for every object we consider its incidence angle to the sensor, as well as its orientation angle in the coordinate system of reference. As shown in the right of the figure, we rotate each object to make it axis-aligned and orient it towards the front of the sensor (i.e., forward with respect to the ego vehicle), then we position it around the sensor such that its incidence angle remains unchanged. Depending on which $g$ of the $G$ positions the rotated axis-aligned and forward-facing object is located, we select the corresponding vector field $g$ to perturb it. This is described in Figure~\ref{fig:relative_rotation} via two examples.

When using ground truth or predicted 3D bounding boxes (Section~\ref{sec:anchor_points}), the object orientation is available. This allows the 12 vector fields to specialize on different instances, such that each can be specific to objects having points located in certain regions of the default bounding box $B_0$. This means applying each vector field $g$ only to the objects appearing in its corresponding slice around $\beta_g$, as shown in the figure. However, without oriented 3D bounding boxes, but using axis-aligned boxes determined by the point-level annotations (Section~\ref{sec:anchor_points}), it is not possible to establish the orientation of the objects. Therefore, with $G=12$, opposite pairs of vector fields would overlap, e.g., 1 and 7 in Figure~\ref{fig:relative_rotation}. To avoid this issue, with axis-aligned boxes we consider only $G=6$ vector fields (1 to 6), and use them at both their nominal positions and the corresponding opposite ones.

\subsection{Adversarial Data Augmentation}
\label{sec:augmentation}
After learning to deform objects in the first stage (Figure~\ref{fig:framework}), we use the vector field as adversarial augmentation. In this phase, we train a model for a downstream task (e.g., 3D object detection or 3D semantic segmentation) and use the adversarial vectors to perturb the scenes in input as data augmentation. This increases the robustness of the model and also its generalization capability, as we expand the training data in a plausible way thanks to our combination of adversarial examples and constraints (e.g., sliding points only along their sensor view ray). Since the learned perturbations are structurally-consistent, they are better suited than standard augmentations (e.g., scaling, flip, rotation) to resemble rare out-of-domain object shapes, such as cars from a different country~\cite{wang2020train}.

Instead of applying the same set of vectors over the entire dataset, we increase the diversity of the training data by using $N$ different vector fields for each of the $G$ rotations (Section~\ref{sec:field_application}). These $N \times G$ vector fields are learned independently from one another and they are randomly initialized.
During training of the model for the downstream task, we randomly select one object in the scene and we deform it with a randomly chosen vector field out of the $N$ available ones for its relative rotation.

We do not augment all instances in the scene to let the model learn unperturbed objects as well as deformed ones. Moreover, thanks to the $N$ different sets of vectors, the same object can be deformed differently at different epochs as different vectors are chosen. Furthermore, the same vector field does not deform all objects the same way, as it gets scaled and rotated to match the object it is applied on. Then, since the 3D points do not lie on the same grid as the vectors, their shift depends on how close they are from which vector. By considering neighboring vectors and the relative distances of each point to the closest vectors, even slightly different 3D locations of the points result in different deformations.

Overall, we introduce a wide variability and randomness in the training process which aid generalization and prevent overfitting to specific deformations, as demonstrated in the next section through extensive experiments.



\section{Experimental Setup}

\subsection{Datasets}
We conducted our experiments on seven different datasets. Six of them are autonomous driving datasets with available LiDAR data: KITTI~\cite{geiger2012we}, the Waymo Open Dataset both for semantic segmentation and object detection~\cite{sun2020waymo}, our proposed synthetic CrashD, SemanticKITTI~\cite{behley2019semantickitti}, and nuScenes~\cite{caesar2020nuscenes}. Additionally, in our previous conference publication we applied our method also on the indoor SUN RGB-D dataset~\cite{song2015sun}, to demonstrate the wide applicability of our approach, also to time-of-flight (ToF) cameras.
All datasets used in this work are openly available, provided by the respective authors cited in the reference section.

\subsubsection{3D Object Detection Datasets}
We report on four different datasets for 3D object detection.
\textbf{KITTI}~\cite{geiger2012we} is a popular 3D object detection benchmark recorded in Germany. We adopted a standard split~\cite{lang_pointpillars_2019}, which comprises 3712 training and 3769 validation LiDAR point clouds, where we used the \textit{car} class, reporting on the standard \textit{easy}, \textit{moderate} and \textit{hard}.
We evaluated models trained on KITTI for 3D object detection on Waymo and our CrashD (without any fine-tuning) to assess the generalization capability of the models to out-of-domain data, particularly critical for autonomous driving.
The \textbf{Waymo} Open Dataset~\cite{sun2020waymo} is a challenging large-scale collection of real scenes recorded in various locations of the USA. It is highly diverse with different weather and illumination conditions, such as rain and night. Specifically, we evaluated on the official validation set.
For LiDAR 3D object detection, we trained solely on KITTI, transferred the models to Waymo and CrashD, and considered only the \textit{car} class.

\textbf{Proposed CrashD dataset}
To estimate the generalizability of a model on rare out-of-domain data, in our previous conference publication we openly released a synthetic dataset, which we called CrashD (\cite{lehner20223dvfield}). This dataset was designed as out-of-domain test benchmark for 3D object detectors trained on KITTI~\cite{geiger2012we}, Waymo~\cite{sun2020waymo} or similar datasets.
The proposed CrashD includes various types of rare cars with different shapes, such as old, sports, and damaged. While if included at all these categories are normally only long tail samples in standard datasets, in CrashD half of the cars are damaged, and half have an unusual shape (e.g., classic cars). The other half of each category is undamaged, or normal, serving as control group to assess the performance gap of a model between standard cars and rare out-of-domain ones.
Specifically, the crashes were individually generated with a realistic physics simulator~\cite{BeamNGTechnicalPaper21}. We generated various types of crashes and we distinguished them depending on the intensity, namely \textit{light}, \textit{moderate}, \textit{hard}, as well as the kind of damage: \textit{clean} (i.e., undamaged), \textit{linear} (i.e., frontal or rear), and \textit{t-bone} (i.e., lateral).
We randomly and automatically generated 15340 scenes and captured them with a 64-beam LiDAR, which we configured to mimic the one in KITTI. Each scene features between 1 and 5 cars, with visible damages (\textit{crash} set). Then we collected the \textit{clean} set after repairing the vehicles and placing them at the same locations. Overall, CrashD contains 46936 cars. CrashD is available for download through the dataset website \hyperref[{sec:crashd_url}]{$^\Diamond$}.
Further details can be found in our conference publication and its supplementary material~\cite{lehner20223dvfield}.
\blfootnote{$^\Diamond$ \url{https://crashd-cars.github.io/}\label{sec:crashd_url}}

\subsubsection{3D Semantic Segmentation Datasets}
We report results for 3D semantic segmentation on three different autonomous driving datasets.
\textbf{SemanticKITTI} \cite{behley2019semantickitti} is a popular 3D semantic segmentation benchmark recorded in Germany.
Behley et al.~annotated with semantic classes the LiDAR point clouds from the KITTI odometry task. SemanticKITTI includes point-level annotations for 23201 full 360° scans for training (including validation) and 20351 for testing. We followed the convention, using sequence 08 as validation set (i.e., 4071 samples), and the remaining annotated sequences for training (i.e., 19130 samples). We evaluated the models across the standard 19 classes, 8 of which are objects.
As for 3D object detection, we are interested in assessing the generalization and robustness of the models, so we include two additional datasets as transfers (without any fine-tuning).
The \textbf{Waymo} Open Dataset~\cite{sun2020waymo} was recently extended with point-level annotations for a subset of the 3D object detection scans. The 3D semantic segmentation dataset includes labels for 23 different classes.
\textbf{nuScenes}~\cite{caesar2020nuscenes} is another large scale autonomous driving dataset. It contains around 15h of driving data collected in Boston and Singapore, with diverse traffic scenarios (e.g., both left and right hand drive). Similarly to Waymo, it includes more challenging weather and lighting conditions (e.g., rain and night) compared to SemanticKITTI, making it a difficult out-of-domain test for our models.
For Waymo and nuScenes, we evaluated on their validation sets.

\subsubsection{Domain Generalization}
As we did not train our models on Waymo, nuScenes or CrashD, but only used these as generalization tests after training solely on KITTI (detection) SemanticKITTI (segmentation), various challenges arised when transferring.
KITTI and SemanticKITTI were captured with the same sensors and we designed CrashD to mimic the LiDAR from KITTI. However, the sensors used in the other transfer datasets are rather different, thereby providing highly different 3D point clouds. Scans from Waymo are 50\% denser than those of KITTI, and the field of view is narrower on Waymo.
Instead, nuScenes was captured with a 32-beam LiDAR, delivering significantly sparser point clouds than those captured with the 64-beam sensor of SemanticKITTI.
For object detection, we report on the proposed CrashD to show the model performance at the long tail of the data distribution, featuring rare out-of-domain samples, such as damaged and rare cars. Instead, Waymo is used to demonstrate the capabilities on challenging real-world data from a different country captured by a different 3D sensor.
For the same reasons, for semantic segmentation, we conducted experiments on Waymo and nuScenes. However, we used both in order to assess the models performance with denser and sparser point clouds compared to those used for training.
Overall, the difficult set of transfers included in this work demonstrates the ability of the models to generalize beyond their training specifications.

\textbf{Semantic classes for transfers} Since the annotation specifications adopted for each dataset are different, we had to map the classes accordingly. This was not an issue for object detection (\cite{lehner20223dvfield}) as we focused on the \textit{car} class, which is available on all datasets used. Instead, for semantic segmentation, similarly to previous works~\cite{yi2021complete_and_label}, we could not evaluate on all classes. In fact, segments such as \textit{road} were annotated differently and had to be ignored for the transfer: Waymo~\cite{sun2020waymo} included also the driveways connecting parking lots with the road over sections of sidewalks, while SemantiKITTI did not. Nevertheless, on SemanticKITTI all our models (including Cylinder3D) were trained on the 19 classes available. When transferring we considered only the compatible classes: 12 for Waymo and 8 for nuScenes. Instead, to compare with~\cite{yi2021complete_and_label}, we used their class mapping: 2 classes for Waymo and 10 for nuScenes. The details of the mappings can be found in our Supplementary Material.

\subsubsection{Annotation Requirements}
As described in Section~\ref{sec:anchor_points}, it is crucial to associate the adversarial vectors to a region of the point cloud. We did this via bounding boxes. Therefore, for object detection training our method does not require any extra annotations, other than the 3D bounding boxes which are readily available for training a detector. Instead, a semantic segmentation dataset does not necessarily include bounding box annotations. We circumvent this in 2 alternative ways: via an off-the-shelf 3D object detector, or via axis-aligned bounding boxes. The former strategy implies that the detector has been trained with 3D bounding box annotations on a different dataset. The latter requires point-level annotations of instances, which may not be available. Exploring both strategies ensures the applicability of our method. Conversely, at inference time our adversarially augmented models do not have any requirements in terms of annotations, making them as applicable as standard models.

\subsection{Evaluation metrics}
We evaluated the 3D object detection performance on the standard \textbf{AP}, with a 3D IoU threshold of 0.7 for KITTI and CrashD, 0.5 for Waymo. To measure the quality of the adversarial perturbations for object detection we followed \cite{tu_physically_2020} using the attack success rate (\textbf{ASR}) metric. It measures the percentage of objects that become false negatives after undergoing an adversarial alteration.
For the ASR, we considered an object detected if its 3D IoU was larger than 0.7.
Instead, for 3D semantic segmentation, we evaluated on the common \textbf{mIoU} as the mean over the classes IoUs. For the adversarial attacks, we computed the IoU on the adversarial examples generated from the validation set.

\subsection{Network architectures}
For 3D semantic segmentation we used Cylinder3D~\cite{zhu2021cylindrical}, which divides the point cloud into voxels with distance-dependent sizes.
For 3D object detection, we used three different detectors.
PointPillars~\cite{lang_pointpillars_2019} voxelizes the scene in vertical columns (i.e., pillars) from the bird's eye view, using PointNet~\cite{qi_pointnet_2017} for feature extraction. Part-A$^2$ Net~\cite{shi2019parta2} is an extension of PointRCNN that predicts intra-object part locations for improved accuracy. VoteNet~\cite{qi2019_votenet} is based on PointNet++ and Hough voting. While PointPillars and Part-A$^2$ Net are mostly used for autonomous driving settings, VoteNet is used indoor.

\subsection{Implementation details}
For the \textit{car} class we constructed each vector field within $B_o$ with $w=1.8$m, $h=1.6$m, $l=4.6$m and a step size of $t = 20$cm resulting in 1656 vectors per vector field. If not stated otherwise, we grouped objects by relative rotations with $G = 12$ groups, and set $N = 6$. For the \textit{person} class we used axis-aligned boxes with the dimensions $w=0.54$m, $h=1.7$m, $l=0.66$m and a step size of $t = 5$cm resulting in 5036 vectors per vector field. Due to the ambiguous direction of the axis-aligned boxes, we only use $G = 6$ vector fields, each at its position and the opposite one with respect to the sensor.
During the perturbation stage, we moved points according to their $k=2$ nearest vectors and deformed only along the sensor ray.
For the PGD optimization, we used Adam with a learning rate of 0.05 for object detection and 0.01 for semantic segmentation. The distance threshold was set to $\epsilon = 30$cm and $\psi = 0.3$ for the intensity. Each vector was randomly initialized form a uniform distribution with values between -1cm and 1cm.

For the second stage (i.e., adversarial augmentation, training of the 3D object detection or the 3D semantic segmentation model), we used all the same hyperparameters and configurations provided by the authors of each model, apart from adding our adversarial examples at training time (Section~\ref{sec:field_application}).
We trained all models using PyTorch. For 3D object detection we used the MMDetection3D framework~\cite{mmdet3d2020}, while for 3D semantic segmentation we used the code provided by~\cite{zhu2021cylindrical}. All models were trained on a single NVIDIA Tesla V100 32GB GPU.

While for object detection we did not use the LiDAR intensity signal, all segmentation models took it in input, unless otherwise noted (e.g., Table~\ref{table:intensity_lidar}).
Specifically, the intensity values of SemanticKITTI are provided by the authors in the range $[0,1]$. Conversely, Waymo intensity values are unbounded, and those of nuScenes are in the range $[0,255]$. Since we trained all semantic models on SemanticKITTI, we scaled the intensities of the other datasets to match the same range. For Waymo, we followed \cite{zhu2021cylindrical} adjusting them with \textit{tanh}. For nuScenes, we normalized them to $[0,1]$.

\subsection{Prior works and Baselines}
We compared our work with other adversarial methods, which we used to generate adversarial examples for augmenting the training data.
For 3D object detection, all models were applied on PointPillars~\cite{lang_pointpillars_2019}, unless otherwise noted. Instead, for 3D semantic segmentation, all were based on the Cylinder3D framework~\cite{zhu2021cylindrical}. We selected Cylinder3D thanks to its strong performance on the popular LiDAR semantic segmentation benchmarks (i.e., SemanticKITTI and nuScenes).

To represent point perturbation methods, we used the iterative gradient L2~\cite{xiang_generating_2019} and the Chamfer attack~\cite{liu2020adversarial}. For adversarial generation, we used \cite{xiang_generating_2019} adding 10\%, while for removal we used \cite{yang2019adversarial_removal} removing 10\% of the object points. We applied these adversarial approaches for both 3D object detection and 3D semantic segmentation.

For a fair comparison, all models were trained on the same dataset split: for object detection we used the KITTI split used by~\cite{lang_pointpillars_2019}; for semantic segmentation we used the standard split of SemanticKITTI~\cite{behley2019semantickitti}.
Moreover, for all approaches we used $\epsilon = 30$cm, $\psi = 0.3$ and we altered the point clouds as data augmentation with the same settings as for our approach (i.e., random selection of one object per scene to augment).

For 3D semantic segmentation, we compared ours with the domain adaptation method of~\cite{yi2021complete_and_label}, which was designed to address the disparities between different LiDAR sensors.
Furthermore, for 3D object detection we combined our approach with the domain adaptation statistical normalization (SN) strategy of~\cite{wang2020train}. Following them, after computing the average box dimensions in the target datasets (i.e., Waymo and CrashD), we scaled the source (i.e., KITTI) point clouds within the ground truth boxes accordingly and fine-tuned the trained 3D object detector with this altered target-aware source data.

\section{Quantitative Results}
\label{sec:quantitative}
In this section, first, we discuss the performance of adversarial methods towards out-of-domain generalization (Section~\ref{sec:results_main_generalization}), then adversarial methods as attacks (Section~\ref{sec:results_attacks}) and the trade-off between attack strength and generalization (Section~\ref{sec:specificity-generalization}). Furthermore, we explore the benefits of the LiDAR intensity signal (Section~\ref{sec:results_intensity}), explaining its impact by comparing its distribution across classes and datasets (Section~\ref{sec:results_intensity_distributions}). Then, we assess the robustness of our approach against varying intensity inputs (Section~\ref{sec:results_intensity_robustness}). Moreover, we compare the vectors learned for different tasks and their impact on adversarial augmentation across tasks (Section~\ref{sec:results_vectors_comp}), we compare our approach with domain adaptation (Section~\ref{sec:results_DA_comp}) and combine with it too (Section~\ref{sec:results_DA_combination}), as well as with standard data augmentations (Section~\ref{sec:results_combination_dataaug}). Additionally, we discuss the impact of our vectors across different architectures (Section~\ref{sec:results_3D_detectors}), their benefits in terms of robustness at further distances (Section~\ref{sec:results_distances}), and also the effect of different data annotations (Section~\ref{sec:exp_annotations}). Finally, we discuss the impact of our constraints in an ablation study (Section~\ref{sec:results_ablation_learning}). Additional results can be found in the supplementary material of this work, as well as in our conference publication and its supplementary material~\cite{lehner20223dvfield}.

Throughout the results section, we use consistent IDs denoting trained models and vectors, to help associating different experiments and evaluations. Within each line of a table, all results across different datasets are obtained by the same model trained only on KITTI or SemanticKITTI (depending on the task), and identified by the ID, except for the attack performance, which is always computed on the baseline (i.e., PointPillars or Part-A$^2$ for object detection, and Cylinder3D for semantic segmentation).

\subsection{Adversarial Methods and Generalization}\label{sec:results_main_generalization}

\subsubsection{Semantic Segmentation}
\textbf{Generalization data}
As described by previous works (\cite{wang2020train,gasperini2021r4dyn}), nuScenes and Waymo feature rather different scenarios compared to those of KITTI or SemanticKITTI.
Both transfers are particularly challenging due to the different countries in which the data was captured, leading to different vehicles and street layouts. An additional major challenge is due the different LiDAR sensors used to capture the point clouds. Compared to SemanticKITTI, Waymo has a 50\% higher point density and a narrower field of view~\cite{sun2020waymo}, while nuScenes is significantly sparser, especially at further distances.
Therefore, evaluating the transfer from SemanticKITTI to Waymo and nuScenes (without fine-tuning) assesses the benefits of the adversarial deformations towards the segmentation of previously unseen and out-of-domain real objects and scenes. In these settings, preserving the plausibility of the perturbations is crucial to improve the generalization of a model by expanding its training data distribution without introducing samples that are too far from it. Our method addresses this via plausibility constraints (e.g., sensor-awareness) and adversarial examples.

\setlength{\tabcolsep}{8.2pt}
\begin{sidewaystable}
\begin{center}
\begin{minipage}{\textheight}
\begin{tabular*}{\textheight}{ll|l|cccc|cccc|cccc}
\toprule

\multicolumn{3}{c|}{~} & \multicolumn{4}{c|}{SemanticKITTI} & \multicolumn{4}{c|}{$\rightarrow$ Waymo} 
& \multicolumn{4}{c}{$\rightarrow$ nuScenes}
\\ 

ID & Attack & \multicolumn{1}{l|}{Method} & mIoU & \textit{car} & \textit{pers.} & \textit{bicyc.} & mIoU & \textit{car} & \textit{pers.} & \textit{bicyc.} 
& mIoU & \textit{car} & \textit{pers.} & \textit{bicyc.}
\\

\midrule

c.o & - & Cylinder3D & \textbf{64.39} & 96.02 & 73.84 & 47.15 &   29.43 & 72.10 & 6.69 & 5.68 & 29.97 & 66.04 & 0.00 & 0.89 \\
s.l & a.l* & iterative gradient L2* & 56.30 & \uline{77.22} & 72.45 & 47.39 & 24.86 & \uline{24.48} & 8.93 & 1.07 & 26.15 & \uline{17.93} & 0.12 & 0.25 \\
s.c & a.c* & Chamfer attack* & 58.36 & \uline{83.72} & 64.94 & 47.85 & 27.21 & \uline{35.48} & 20.61 & 4.55 & 22.40 & \uline{12.33} & 0.00 & 0.20 \\
s.g & a.g* & adversarial generation* & 62.94 & \uline{95.51} & \textbf{75.87} & 47.86 &  26.70 & \uline{64.19} & 5.10 & \textbf{8.31} & 28.05 & \uline{66.15} & 0.02 & 0.67 \\
s.r & a.r* & adversarial removal* &  63.43 & \uline{95.86} & 74.27 & 48.56 &  24.29 & \uline{70.73} & 4.44 & 5.45 & 29.48 & \uline{64.22} & 0.05 & 0.73 \\

u.c & $N\times$a.u.c & [ours] untargeted \textit{car} & 63.85 & \uline{96.03} &	77.11 &	33.41 & \textbf{37.74} & \uline{\textbf{82.38}} & 17.13 & 5.04 & \textbf{31.55} & \uline{\textbf{71.92}} & 0.06 & 0.53 \\
t.o & $N\times$a.t.o & [ours] targeted \textit{oth.v.} & 64.15 & \uline{\textbf{96.66}} &	72.88 &	49.21 & 30.39 & \uline{75.70} & 9.88 & 3.66 & 29.57 & \uline{66.65} & 0.00 & 0.00 \\
t.b & $N\times$a.t.b &[ours] targeted \textit{bicycle} & 62.75 & \uline{95.58}	& 74.49	& \uline{\textbf{54.27}} & 33.43 & \uline{76.14} & \textbf{23.12} & \uline{7.11} & 30.06 & \uline{66.64} & 0.00 & \uline{\textbf{1.87}} \\
t.m & $N\times$a.t.m & [ours] targeted \textit{motorc.} & 60.80 & \uline{96.41} & 72.68 & 41.63 & 30.28 & \uline{76.03} & 20.31 & 3.28 & 29.82 & \uline{68.40} & 0.00 & 1.40\\
u.p & $N\times$a.u.p & [ours] untargeted \textit{pers.} & 63.12 & 96.35 & \uline{75.82} & 50.11 & 33.85 & 74.47 & \uline{22.22} & 4.24 & 29.44 & 60.59 & \uline{\textbf{0.17}} & 0.58\\


\bottomrule

\end{tabular*}
\caption{Comparison of \protect\culine{3D semantic segmentation} models trained on SemanticKITTI~\cite{behley2019semantickitti} towards out-of-domain data (without any fine-tuning), namely Waymo~\cite{sun2020waymo} and nuScenes~\cite{caesar2020nuscenes} validation sets. Each method applies a data augmentation, and apart from the baselines (first two lines) all others are adversarial augmentation approaches. All models are based on Cylinder3D~\cite{zhu2021cylindrical}. Underlined values represent the classes that are directly affected by the adversarial attacks. $\rightarrow$: transfer from SemanticKITTI. *: sample-specific. \textit{pers.}: \textit{person}; \textit{bicyc.}: \textit{bicycle}; \textit{oth.v.}: \textit{other-vehicle}; \textit{motorc.}: \textit{motorcycle}.}
\label{table:main_semantic}
\end{minipage}
\end{center}
\end{sidewaystable}

\begin{sidewaystable}
\setlength{\tabcolsep}{12.35pt}
\begin{center}
\begin{minipage}{\textheight}
\begin{tabular*}{\textheight}{ll|l|ccc|r|c|cc|cc}
\toprule

\multicolumn{3}{c|}{~} & \multicolumn{4}{c|}{KITTI} & $\rightarrow$ Waymo & \multicolumn{4}{c}{$\rightarrow$ CrashD} \\

\multicolumn{3}{c|}{~} & \multicolumn{3}{c|}{AP} & & & \multicolumn{2}{c|}{AP \textit{normal}} & \multicolumn{2}{c}{AP \textit{rare}} \\

ID & \multicolumn{1}{l|}{Architecture} & \multicolumn{1}{l|}{Method} & \textit{easy} & \textit{mod.} & \textit{hard} & ASR & AP & \textit{clean} & \textit{crash} & \textit{clean} & ~~\textit{crash} \\

\midrule

p.n & \multirow{10.4}{*}{PointPillars} 
& no augm. & 70.00 & 61.88 & 56.23 & -~~~ & 30.68 & 1.79 & 0.93 & 3.92 & 2.33 \\
p.s & & no obj.~sampl. & 83.83 & 74.14 & 68.30 & -~~~ & 37.85 & 50.36 & 36.44 & 28.70 & 20.02 \\
p.p & & PointPillars &  \textbf{88.24} & 77.11 & 74.55 & -~~~ & 40.86 & 65.20 & 43.67 & 34.14 & 22.48  \\

p.l & & iter.~grad.~L2 * &  86.24 &76.92 & 73.84 & $^*$95.9 & 39.86 & 58.65 & 41.86 & 35.92 & 23.69 \\
p.c & & Chamfer att.* &  87.15 & 77.05 & 74.07 & $^*$\textbf{99.8} & 40.54 & 56.84 & 39.56 & 36.29 & 24.73\\
p.g & & advers.~gener.* & 86.12 & 76.39 & 73.18 & $^*$91.6 & 40.55 & 57.75 & 38.03 & 35.73 & 24.18 \\
p.r & & advers.~remov.* & 86.51 & 76.85 & 74.04 & $^*$86.1 & 40.32 & 66.52 & 48.88 & 41.42 & 28.10 \\
p.o & & [ours] & 87.05 & \textbf{77.13} & \textbf{75.55} & 63.4 & \textbf{44.61} & \textbf{67.95} & \textbf{52.87} & \textbf{43.40} & \textbf{30.37} \\

\cmidrule( r){3-12}

d.a & & SN dom.~adapt. & -&-&-&-~~~ & 49.27 & 79.42 & 72.59 & 60.53 & 48.23 \\
d.a.o & & [ours] + SN & -&-&-&-~~~ & \textbf{51.32} & \textbf{92.14} & \textbf{87.28} & \textbf{86.26} & \textbf{76.42} \\


\cmidrule(lr){1-12}
p.a & \multirow{2}{*}{Part-A$^2$} & Part-A$^2$ & 89.60 & 79.16 & 78.52 & -~~~ & 49.76 & 83.05 & 63.25 & 74.03 & 52.33\\
p.a.o & & [ours] & \textbf{89.65} & \textbf{79.26} & \textbf{78.62} & 50.5 & \textbf{56.08}  & \textbf{88.80} & \textbf{73.80} & \textbf{81.10} & \textbf{61.34}\\

\bottomrule

\end{tabular*}
\caption{Comparison of \protect\culine{3D object detection} methods trained on KITTI~\cite{geiger2012we} towards out-of-domain data (without any fine-tuning), namely Waymo validation set~\cite{sun2020waymo} and our CrashD datasets. Each method applies a different data augmentation (for adversarial ones ASR is measured on their adversarial examples), or performs domain adaptation (only SN~\cite{wang2020train}), resulting in the reported APs. Only the \textit{car} class is considered.
$\rightarrow$: transfer from KITTI. $^*$: sample-specific, so adversarial examples are tailored to the samples being evaluated (i.e., validation set).
}
\label{table:AP_on_all}
\end{minipage}
\end{center}
\end{sidewaystable}

\textbf{Compared methods}
Table~\ref{table:main_semantic} shows the comparison between our approach and related adversarial methods towards out-of-domain generalization for 3D semantic segmentation. All models are applied on Cylinder3D~\cite{zhu2021cylindrical}. In particular, we report other adversarial perturbation methods used as adversarial augmentations, namely the iterative gradient L2~\cite{xiang_generating_2019} and the Chamfer attack~\cite{liu2020adversarial}, adversarial generation~\cite{xiang_generating_2019}, as well as adversarial removal~\cite{yang2019adversarial_removal}.

\textbf{Results}
Although none of the adversarial augmentations improved the in-domain mIoU (i.e., on SemanticKITTI), our approach was the only one to improve upon the baseline on the transfers to both Waymo and nuScenes.
Our method applied on the \textit{car} class in an untargeted fashion (u.c) reached the highest mIoU and IoU for \textit{car}, in both Waymo and nuScenes.
Specifically, on Waymo, the IoU of our untargeted approach was higher than that of the standard Cylinder3D (c.o) by over 10 (+14\%) for \textit{car} and over 8 (+28\%) on average on all classes (mIoU). On the sparser point clouds of nuScenes, it was higher by 6 (+9\%) on car and 1.6 (+5.3\%) on average.
This shows the effectiveness of our adversarial augmentations to significantly improve the robustness against challenging out-of-domain data.

The IoU scores obtained by the standard Cylinder3D on nuScenes~\cite{caesar2020nuscenes} were particularly low. For \textit{person}, the standard Cylinder3D scored 0.00. This can be attributed to the highly sparse point clouds of nuScenes, which did not have enough points on smaller classes (e.g., \textit{person} and \textit{bicyle}) to be accurately processed by Cylinder3D trained on the denser SemanticKITTI. An additional reason is the different LiDAR intensity distributions (Section~\ref{sec:results_intensity_distributions}). Since the architecture on which every method in the table is based upon (i.e., Cylinder3D) performed so poorly on nuScenes, the performance gains on this dataset were not as significant as on Waymo. Nevertheless, our untargeted \textit{car} configuration (u.c) outperformed the standard Cylinder3D and all other adversarial approaches for both mIoU and the reference class \textit{car}.

Sample-specific perturbation approaches (Chamfer and the iterative gradient L2 attacks), by introducing severely altered samples, which are too far from the existing training data, worsened the generalization performance when used for adversarial augmentation. Instead, adversarial generation and removal are less disruptive, by acting upon fewer points (i.e., 10\%). Nevertheless, unlike our method, they also worsened the IoU on the reference class \textit{car}.

\textbf{Targeted adversarial augmentations}
While our untargeted augmentations (u.c) performed better overall, especially in the transfers to Waymo and nuScenes, targeted ones can be deployed to strengthen a specific class boundary. Depending on the specific use cases, errors may be valued differently for different classes. For an autonomous vehicle, confusing \textit{vegetation} with \textit{trunk} may not be as severe as confusing \textit{car} with \textit{road}, or \textit{car} with \textit{bicycle}. This is linked with the impact that such errors have on downstream tasks (e.g., trajectory prediction and path planning). Therefore, it may be of interest to strengthen specific class boundaries of a model to avoid severe confusions (e.g., \textit{car}-\textit{bicycle}), even at the cost of weakening a different boundary (e.g., \textit{car}-\textit{other-vehicle}).
As shown in Table~\ref{table:main_semantic} our targeted adversarial augmentations offer this valuable option. In the table we report various augmentations based on targeted adversarial attacks on the points of the \textit{car} class towards the corresponding targeted class (e.g., \textit{bicycle}). Among our models, augmenting \textit{car} points while targeting \textit{bicycle} led to the highest IoUs for the \textit{bicycle} class, both in- and out-of-domain, showing the impact of our targeted techniques to alter specific decision boundaries of the model. The other targeted augmentations acted on different decision boundaries, depending on their target class. For example, targeting \textit{other-vehicle} improved the in-domain IoU for \textit{other-vehicle} by 9.3 points over the standard Cylinder3D, reaching 63.8.

\textbf{Class \textit{person}}
Table~\ref{table:main_semantic} reports also our untargeted adversarial augmentations for the class \textit{person}. Given that this class has significantly less points than \textit{car} (477K compared to 30.8M on the validation set of SemanticKITTI), both our adversarial vectors and our augmentations operate on a substantially smaller set, which can hinder the effectiveness, compared to \textit{car}. Nevertheless, despite the reduced data available, our adversarial augmentations improved the generalization both in-domain and out-of-domain (Waymo), compared to the standard Cylinder3D. Especially for the \textit{person} class, the IoU improved by 2 points on SemanticKITTI, and increased by a substantial 3.3x on Waymo. These results show the effectiveness of our approach on a class different than \textit{car}, while operating without 3D bounding boxes.

\subsubsection{Object detection}
\textbf{Generalization data}
Similarly to Table~\ref{table:main_semantic} for 3D semantic segmentation, Table~\ref{table:AP_on_all} compares our method with other adversarial approaches for 3D object detection. Here, we apply ours and baseline approaches on PointPillars~\cite{lang_pointpillars_2019}. As for 3D semantic segmentation, none of the adversarial approaches improved the in-domain AP on KITTI compared to the standard PointPillars. However, our adversarial augmentations significantly improved the performance on out-of-domain data. As previously shown by~\cite{wang2020train}, the transfer from KITTI to Waymo is particularly difficult due to the different shapes and sizes of the vehicles, which depend on the country where the data was collected (i.e., Germany and USA, respectively), as well as the different street layouts and urban landscapes. Moreover, the Waymo point clouds are 50\% higher in density and its LiDAR sensor has a narrower field of view~\cite{sun2020waymo}.
Depending on the global markets, the distribution of vehicle types varies significantly. This is due to certain manufacturers selling different models in different markets, as well as their market penetration, and the preferences and needs of the local population.
Therefore, this challenging transfer evaluates the quality of the adversarial perturbations compared to the real vehicles found in different countries.

\setlength{\tabcolsep}{7.5pt}
\begin{table*}[t]
\begin{center}
\begin{tabular}{lll|l|ccccccc}
\toprule


ID & Ad.Cl. & Tar.Cl. & Adversarial Attack & mIoU & \textit{car} & \textit{pers.} & \textit{bicyc.} & \textit{oth.v.} & \textit{motorc.} & \textit{truck} \\
\midrule

(c.o) & - & - & (Cylinder3D) & 64.39 & 96.02 & 73.84 & 47.15 & 54.42 & 69.11 & 88.96 \\
\cmidrule(lr){1-11}
a.l* & \multirow{4}{*}{\textit{car}} & any & 
iterative grad.~L2 * & \textbf{41.16} & \textbf{11.18} & 70.30 & \textbf{18.88} & 38.10 & 47.10  & \textbf{3.30} \\
a.c* & & any & Chamfer attack * & 50.92 & 17.13 & 73.10 & 46.56 & 50.70 & 52.94  & 25.88 \\
a.g* & & any & advers.~generation * &  64.06 & 95.73 & 73.81 & 46.96 & 53.94 & 67.97 & 88.23 \\
a.r* & & any & advers.~removal * &  64.38 & 95.71 & 73.86 & 47.13 & 54.56 & 69.02 & 89.11 \\
\cmidrule(lr){1-11}
a.u.c\label{u.c} & \multirow{7}{*}{\textit{car}} & any & [ours] untargeted &  53.52 & \protect\culine{62.59} & 68.98 & 36.80 & 16.79 & 46.47 & 49.22 \\
a.u.- &  & any & [ours] untarg.-10\% & 54.65 & \protect\culine{66.01} & 68.06 & 37.86 & 18.09 & 52.02 & 55.25  \\
a.u.a &  & any & [ours] untar.ax-alg. & 61.29 & \protect\culine{89.02} & 71.43 & 46.16 & 38.58 & 65.19 & 80.38  \\
a.t.b & & \textit{bicyc.} & [ours] tar. \textit{bicyc.} & 53.89 & 68.74 & 67.11 & \protect\culine{33.11} & 20.20 & 45.08 & 52.07 \\
a.t.o & & \textit{oth.v.} & [ours] tar. \textit{oth.v.} &  54.61 & 65.91 & 68.96 & 38.10 & \protect\culine{\textbf{16.43}} & 52.04 & 53.88 \\
a.t.m & & \textit{motorc.} & [ours] tar. \textit{motorc.} &  55.62 & 75.79 & 68.40 & 42.07 & 25.25 & \protect\culine{\textbf{27.46}} & 61.20 \\
a.t.t & & \textit{truck} & [ours] tar. \textit{truck} &  55.01 & 76.19 & 71.19 & 39.70 & 22.89 & 49.62 & \protect\culine{40.39} \\

\cmidrule(lr){1-11}

a.u.p & \textit{pers.} & any & [ours] untargeted & 60.68 & 96.01 & \protect\culine{\textbf{30.92}} & 46.07 & 54.50 & 68.55 & 89.00 \\

\bottomrule

\end{tabular}
\caption{
Efficacy of adversarial methods as attacks on \protect\culine{3D semantic segmentation}. All IoUs are computed on the predictions of Cylinder3D (c.o) given deformed point clouds of the validation set of SemanticKITTI by means of each adversarial method. Therefore, a lower IoU translates in a more effective adversarial attack. All methods attack both the 3D location of the points and their intensity values. The adversarial class (Ad.Cl.) indicates the class of which points were altered (e.g., \textit{car}). The target class (Tar.Cl.) shows the class that the adversarial methods aimed to switch the predictions of the adversarial class to. Any as Tar.Cl. from \textit{car} (Ad.Cl.) means altering \textit{car} points in an untargeted fashion, while having \textit{truck} as target means perturbing the point clouds such that the model predicts \textit{truck} for \textit{car} points. *: by being sample-specific, the attack had to be tuned on the same samples on which the method is evaluated (i.e., validation set). Underlined numbers highlight the targeted classes, or the untargeted ones.}
\label{table:attacks_semantic}
\end{center}
\end{table*}

\textbf{Results}
On Waymo, while all other adversarial approaches underperformed the standard PointPillars, our method outperformed it by over 9\%.
Moreover, our adversarial augmentations brought a 13\% improvement on Part-A$^2$~\cite{shi2019parta2}. This shows the effectiveness of our approach towards challenging out-of-domain data also for 3D object detection, thanks to the combination of hard examples with plausible deformations.
The right columns of Table~\ref{table:AP_on_all} are dedicated to the results on the proposed CrashD dataset. Despite transferring the models from KITTI, the AP on \textit{clean normal} cars remained relatively high. This can be attributed to the simplicity of such samples. However, the detection performance dropped significantly when the exact same cars at the same locations were damaged (\textit{crash}).

This perfomance gap shows how far damaged cars are from the training distribution of KITTI, making them natural adversarial examples. The gap increased even further with \textit{rare} cars (i.e., old and sports cars), highlighting again the gap from the source domain (i.e., KITTI) and \textit{normal} vehicles. For these reasons, the most difficult samples of CrashD were those combining both out-of-domain aspects (i.e., rarity and damage) into the \textit{rare crash} group. The AP of PointPillars on this group reduced relatively from \textit{normal clean} by 66\%.

Despite the challenges, our approach significantly outperformed on all transfers and categories the standard object detectors (i.e., PointPillars and Part-A$^2$), as well as all other adversarial methods. Removing points~\cite{yang2019adversarial_removal} was the only adversarial method among the baselines which improved the generalization on CrashD. This could be attributed to its preservation of the overall point clouds, which is an aspect in common with our approach. Nevertheless, removing points adversarially did not improve the generalization to Waymo, which contains denser point clouds and more challenging real scenes. It also did not improve for 3D semantic segmentation.

Similarly to 3D semantic segmentation, the improved generalization can be attributed to the effectiveness of our adversarial augmentations to expand the training data distribution with difficult and plausible examples. Thanks to the added diversity, our method managed to mitigate the domain gap, as shown by the results in the tables.
Overall, the results across Tables~\ref{table:main_semantic} and~\ref{table:AP_on_all} demonstrate the effectiveness of our adversarial augmentations on two highly different tasks, namely 3D semantic segmentation and 3D object detection.

\subsection{Adversarial Methods as Attacks}\label{sec:results_attacks}
In Table~\ref{table:attacks_semantic} we compare the effectiveness of the methods as adversarial attacks for 3D semantic segmentation, in terms of the change in IoU when using their adversarial examples as input for Cylinder3D compared to the untouched inputs.

\subsubsection{Semantic Segmentation, Untargeted Attacks}
Applied on cars, the iterative gradient L2~\cite{xiang_generating_2019} and the Chamfer attack~\cite{liu2020adversarial} were able to majorly reduce the IoU on \textit{car} (Table~\ref{table:attacks_semantic}). Being sample-specific, their deformations had to be learned directly on the data on which they are applied, and in this case also evaluated (i.e., validation set of SemanticKITTI). The same holds true for adversarial generation~\cite{xiang_generating_2019} and removal~\cite{yang2019adversarial_removal}, except that they were not effective attacks as they could not reduce the \textit{car} IoU. This is due to semantic segmentation being a dense task, requiring a prediction for each point. Therefore, removing or adding a few critical points is not enough to lower the IoU as all points count equally towards the metric. In comparison, considering 3D object detection, few points can have a larger impact as they can shift the predicted bounding box (Table~\ref{table:AP_on_all}), thereby reducing the IoU with the ground truth box until it is below the threshold for the AP (e.g., 0.7).
Being sample-independent and learned on the training set, the proposed method is not an attack as effective as the iterative gradient L2 or the Chamfer attacks at reducing the \textit{car} IoU. Nevertheless, our untargeted attack (a.u.c) lowered the \textit{car} IoU by over 33 points. In fact, the proposed method is not meant to render the objects unrecognizable. Instead, we aim to perturb them while preserving their overall shape. As shown in our conference publication~\cite{lehner20223dvfield}, strong attacks do not lead to improved generalization, which is the aim of this work.
Furthermore, in Table~\ref{table:attacks_semantic} we also report the results of our untargeted attack on the \textit{person} class (a.u.p). This effectively lowered the IoU on \textit{person} by 43 points and did not change the \textit{car} IoU with respect to Cylinder3D (c.o). Comparing the \textit{person} attack with the untargeted attack on \textit{car} (a.u.c) shows the effectiveness of our approach, also without using 3D bounding boxes (Section~\ref{sec:axis_aligned}).

\subsubsection{Semantic Segmentation, Targeted Attacks}
In this work, we explored also targeted adversarial attacks and augmentations.
The purpose of the targeted attacks in this context is to test a specific class boundary, rather than limiting the attack to the weakest boundary (i.e., untargeted). We then strengthen this boundary with our targeted adversarial augmentations.
With reference to Table~\ref{table:attacks_semantic}, all our targeted attacks not only correctly reduced the \textit{car} IoU, but also that of their respective targeted class (e.g., \textit{motorcycle}). Specifically, each targeted attack reached the lowest IoU in its class (underlined) among all our attacks, showing their effectiveness. Our untargeted attack often aimed at confusing \textit{car} with \textit{other-vehicle} and \textit{motorcycle}, both decreasing significantly, while not as much with \textit{person}. Therefore, the IoU gap between targeting \textit{other-vehicle} (a.t.o) and not targeting any specific class (untargeted) is rather small for \textit{other-vehicle}. This can be attributed to the \textit{car}-\textit{other-vehicle} decision boundary being weaker than others for the Cylinder3D model examined. However, comparing the performance of Cylinder3D on the standard data (c.o), the perturbed point clouds untargeted (a.u.c) and the targeted ones on \textit{bicycle}, \textit{motorcycle}, or \textit{truck} shows a clear difference on the targeted classes, especially for \textit{motorcycle}.

\subsubsection{Object Detection, Untargeted Attacks}
As seen for semantic segmentation, also for object detection our approach is not an attack as strong as the sample-specific ones. This is shown in terms of ASR in Table~\ref{table:AP_on_all}, where we compared our sample-independent vector fields to their sample-specific point-to-point deformations. As for semantic segmentation, their perturbations were trained directly on the validation set of KITTI, on which the ASR was evaluated. Nevertheless, a very high score on ASR means that the objects became unrecognizable. This goes against aiding generalization, which is the goal of our method.

\subsubsection{Considerations on Attack Efficacy vs.~Generalization}
Unlike adversarial approaches designed as attacks, we do not aim to have the model fully miss our adversarial examples (high ASR). Instead, we want to deform them to increase the robustness on out-of-domain data. Therefore, the adversarial examples need to be altered enough to expand the training data distribution, but not too far from it to prevent confusion for the model. We balance these aspects via sample-independent adversarial perturbations and constraining the deformations. The adversarial attack generates hard examples, while the sample-independence and the constraints (e.g., sensor-awareness) mitigate their strength and preserve their plausibility. As demonstrated throughout this work, this combination is effective to improve generalization and robustness.

\subsubsection{Impact of Data Annotations}
In Table~\ref{table:attacks_semantic}, we explore the impact of the availability of data annotations on our adversarial attack. We explore two different strategies: an off-the-shelf 3D object detector deployed on the training data, and point-level instance annotations. In the latter case, we create axis-aligned bounding boxes around the 3D points constituting an instance. The axis-aligned one prevents the vector fields to specialize on certain object orientations, since such orientation is unavailable, causing sub-optimal performance (a.u.a) compared to using an off-the-shelf detector (a.u.c). In the table, we also assess the effect of the quality of the detector's predictions on our attack. We do so by randomly ignoring 10\% of its bounding boxes. As this impacts directly the performance on the \textit{car} class, the attack is not as effective (a.u.-), whilst managing to significantly reduce the IoU on \textit{car} by 30 IoU. In Table~\ref{table:annotations}, we show the impact of this on generalization.

\subsection{Specificity-generalization Trade-off}\label{sec:specificity-generalization}
As seen in Sections~\ref{sec:results_main_generalization} and~\ref{sec:results_attacks}, strong adversarial examples do not necessarily translate into successful adversarial augmentations. This is because if the attack is too strong (e.g., Chamfer), the generated samples are too far from the existing data distribution, which causes the model to learn objects that are not useful towards the task at hand, thereby degrading the performance. Such extreme transformation can be seen in Figure~\ref{fig:qualitative_attack_semantic}. Therefore, there is a trade-off between an attack strength and its benefits to improve generalization via adversarial augmentation. While this trade-off is evident for sample-specific approaches (Sections~\ref{sec:results_main_generalization} and~\ref{sec:results_attacks}), it is less trivial how it can arise in sample-independent settings, such as ours, where the attack strength is inherently mitigated by not being sample-specific. Towards this end, in this section we explore this trade-off in the context of our method, as we purposely make our attack overfit on fewer samples.

\textbf{Object detection} Table~\ref{table:relative_rotations} shows that by varying the amount of considered relative rotations $G$, a trade-off arises between generalization, attack specificity (i.e., attack strength on individual samples by overfitting to the training data), and storage (i.e., amount of vectors). $G=12$ offers a good balance. Instead, taking $G$ to the extreme and having one vector field for each instance of the dataset, ours would become sample-specific, inheriting the weaker generalization capabilities of prior works~\cite{liu2020adversarial,yang2019adversarial_removal}, as shown already at $G=360$. However, while the sample-specific methods needed to be trained on the validation set, allowing for high ASRs (Table~\ref{table:AP_on_all}), our vectors were learned on the training set. So with high $G$, ours overfitted on the training data, which emerges by evaluating on the validation set.
Our augmentation strategy learns only 1656 3D vectors to perturb objects. However, by training with $G=12$ and $N=6$ (p.o, our standard configuration), the amount of vectors increased to 120K. Conversely, the sample-specific iterative gradient L2~\cite{xiang_generating_2019} and the Chamfer~\cite{liu2020adversarial} attacks required 10.9M and 12.6M vectors for training and validation sets respectively. This shows the easy applicability of our approach.

\begin{table}[t]
\begin{center}
\resizebox{\linewidth}{!}{\begin{tabular}{ll|cc|c|r}
\toprule

& & \multicolumn{2}{c|}{KITTI} & $\rightarrow$ Waymo & \\
ID & \# $G$ & ASR $\uparrow$ & \textit{mod.} & AP & \# vectors \\

\midrule

p.u & 1 & 55.08 & \textbf{77.32} & 40.43 & \textbf{10K} \\ 
p.o & 12 & \textbf{63.37} & 77.13 & \textbf{44.61} & 120K \\
p.f & 360 & 44.84 & 77.06 & 40.30 & 3.6M \\
\bottomrule
\end{tabular}}
\end{center}
\caption{
Our adversarial augmentation method for \protect\culine{3D object detection} applied on PointPillars~\cite{lang_pointpillars_2019} trained on KITTI with varying amounts of relative rotations $G$. $\rightarrow$: transfer without fine-tuning.
}
\label{table:relative_rotations}
\end{table}
\setlength{\tabcolsep}{7.3pt}
\begin{sidewaystable}
\begin{center}
\begin{minipage}{\textheight}
\begin{tabular*}{\textheight}{ll|cc|cccc|cccc|cccc}
\toprule

& \multicolumn{1}{c|}{~} & \multicolumn{2}{c|}{SemK attack} & \multicolumn{4}{c|}{SemanticKITTI} & \multicolumn{4}{c|}{$\rightarrow$ Waymo} 
& \multicolumn{4}{c}{$\rightarrow$ nuScenes} 
\\

ID & \multicolumn{1}{l|}{Method} & mIoU & \textit{car} & mIoU & \textit{car} & \textit{pers.} & \textit{bike} & mIoU & \textit{car} & \textit{pers.} & \textit{bike} 
& mIoU & \textit{car} & \textit{pers.} & \textit{bike}
\\

\midrule

c.n & Cylinder3D w/o intensity & - & - & 59.23 & \textbf{96.23} & 68.67 & \textbf{32.61} & 
37.39 & 63.74 & 37.95 & 4.05 &
32.11 & 56.17 & 0.04 & 0.35
\\

u.n & [ours] w/o intensity & 56.63 & 86.03 & \textbf{59.40} & 95.73 & \textbf{70.35} & 26.53 & 
\textbf{40.36} & \textbf{70.70} & \textbf{52.73} & \textbf{7.57} &
\textbf{34.30} & \textbf{63.13} & \textbf{1.02} & \textbf{2.52}
\\

\cmidrule(lr){1-16}

c.o & Cylinder3D w/ intensity & - & - & \textbf{64.39} & 96.02 & 73.84 & \textbf{47.15} & 
29.43 & 72.10 & 6.69 & \textbf{5.68} &
29.97 & 66.04 & 0.00 & \textbf{0.89}
\\

u.c & [ours] w/ intensity & \textbf{53.25} & \textbf{62.61} & 63.85 & \textbf{96.03} &	\textbf{77.11} &	33.41 & 
\textbf{37.74} & \textbf{82.38} & \textbf{17.13} & 5.04 &
\textbf{31.55} & \textbf{71.92} & \textbf{0.06} & 0.53
\\

\bottomrule

\end{tabular*}
\caption{Impact of the LiDAR intensity signals on \protect\culine{3D semantic segmentation} models trained on SemanticKITTI, both in-domain and when transferring (without fine-tuning) on out-of-domain data. All models are based on Cylinder3D~\cite{zhu2021cylindrical}.
}
\label{table:intensity_lidar}
\end{minipage}
\end{center}
\end{sidewaystable}

\subsection{Impact of LiDAR Intensity}\label{sec:results_intensity}
\subsubsection{Effect In-domain}
In Table~\ref{table:intensity_lidar} we assess the impact of using the LiDAR intensity (also called reflectivity) as extra input for each 3D point. Removing the intensity significantly reduced the mIoU and the IoU on certain classes, such as \textit{person} and \textit{bicycle}, but not on \textit{car}. On Waymo, models not using intensity outperformed the ones relying on intensity by up to 8 mIoU points for the baseline Cylinder3D~\cite{zhu2021cylindrical}. This gap can be attributed to the different sensors used to captured SemanticKITTI and Waymo, which implies different reflectivity measurements. Therefore, for this transfer, not taking the intensity as extra input is advantageous. Nevertheless, our adversarial augmentation significantly improved the predictions both with or without intensity, especially on \textit{car} and \textit{person}.

\subsubsection{Effect on Generalization and Robustness}
A trade-off exists between in- and out-of-domain which depends on the use of the LiDAR intensity values (Table~\ref{table:intensity_lidar}). Since the intensity is a valuable input signal, the in-domain mIoU is higher when integrating it. However, it is better to avoid using it when transferring to different sensors.
Most 3D semantic segmentation experiments in this work used the intensity, because using it maximizes the outcome on the data distribution which is available at the time of development (i.e., in-domain, SemanticKITTI). Instead, we consider the transfer datasets (i.e., out-of-domain, Waymo, nuScenes) as real-world test scenarios, which include data that is not available during the development and training of the method. In realistic scenarios it is not always known on which exact data a model will be applied on.

While different intensity values would arise only by deploying the model on data captured by a different LiDAR sensor, it is possible for example that a trained model gets deployed on data from the fleet of vehicles of a newer generation which includes a different sensor configuration. Therefore, robustness against varying intensity values would mitigate the domain gap introduced by the different sensor and potentially avoid to collect the data again with the newer sensor, simplifying the process.

By generating adversarial examples which alter not only the 3D points (\cite{lehner20223dvfield,xiang_generating_2019,liu2020adversarial,tu_physically_2020}), but also the LiDAR intensity values, our adversarial augmentation approach increases the robustness of the model and makes it less reliant on it. As shown in Table~\ref{table:intensity_lidar}, adding the intensity does not degrade the IoUs of ours as much as for the standard Cylinder3D, thereby effectively mitigating the sensor change.

\subsubsection{Effect on the Attacks}
Comparing the attack performances with and without the intensity (SemK attack in Table~\ref{table:intensity_lidar}), confirms that the LiDAR intensity plays a major role for Cylinder3D. On the reference \textit{car} class, the purely geometrical attack without intensity generated adversarial examples which were not particularly challenging for the model, as shown by the relatively high IoU on \textit{car} after the attack (SemK attack). Nevertheless, using the geometrical 3D vectors that generated these examples for augmentations significantly improved the performance on out-of-domain data. On Waymo, our approach increased the \textit{car} IoU by 11\% without intensity (3D vectors), and 14\% with intensity (4D vectors). On nuScenes, by 12\% without, and 8\% with intensity. In fact, as described in Section~\ref{sec:results_main_generalization}, weaker attacks (e.g., without intensity) can still lead to valuable augmentations.

\begin{figure*}[t]
\centering
  \includegraphics[width=1.0\textwidth]{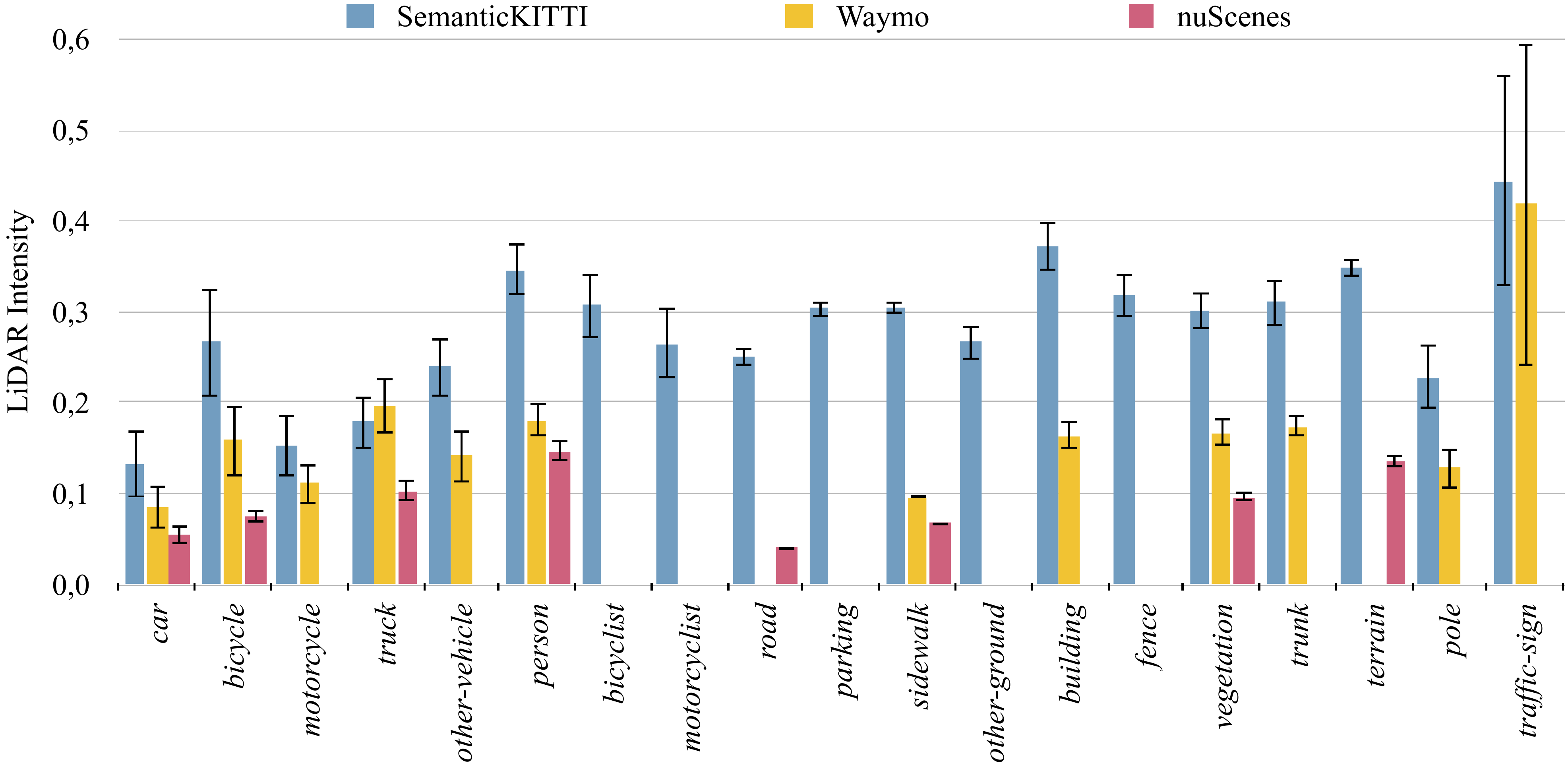}
   \caption{Plot of the mean LiDAR intensity values for each semantic class (x-axis) across the validation sets of the three semantic datasets used, namely SemanticKITTI~\cite{behley2019semantickitti}, Waymo~\cite{sun2020waymo}, and nuScenes~\cite{caesar2020nuscenes}. Each error bar represents the standard deviation within that class and dataset. The classes naming and definition are aligned with those of SemanticKITTI (i.e., 19 classes). Not all classes are represented for the other datasets, due to the non-overlapping definitions with SemanticKITTI (e.g., \textit{building} has no dedicated class in nuScenes). Therefore, the naming follows the convention of SemanticKITTI.}
   \label{fig:intensity_plot}
\end{figure*}

\subsection{LiDAR Intensity Distributions}\label{sec:results_intensity_distributions}
Comparing the distributions of the intensity values for each class across the three datasets shows a significant difference. We report this in Figure~\ref{fig:intensity_plot}.

\subsubsection{Benefits}
First, by looking at the intensity values of SemanticKITTI (Figure~\ref{fig:intensity_plot}), it is evident how useful the intensity signal can be to identify the semantic classes. For SemanticKITTI, the distribution of the class \textit{car} is significantly different from that of \textit{building}, making it easier to separate the two classes. Nevertheless, it should be noted that the LiDAR intensity changes significantly depending on the distance. For SemanticKITTI, between 50 and 60 meters, the mean intensity of \textit{building} is the same as the overall mean of \textit{car}.

\subsubsection{Different Sensors}
While for the most part in SemanticKITTI increasing the distance causes a reduction of the LiDAR intensity values (Figure~\ref{fig:intensity_plot}), this is not the case for Waymo and nuScenes, where the reflectivity is larger at higher distances. This, together with the misaligned distributions shown in Figure~\ref{fig:intensity_plot}, makes it even more challenging to transfer from SemanticKITTI to the other datasets when taking the intensity as input.

\subsubsection{Different Ways to Treat the Values}
Only Waymo provides the raw intensity signals, while SemanticKITTI and nuScenes provide them already scaled, within [0,1] and [0,255] respectively. It is unclear how the scaled values were obtained for SemanticKITTI and nuScenes. To mitigate the impact of the pre-processing of each dataset, we further scale the nuScenes intensity values to [0,1], and apply \textit{tanh} to those of Waymo to constrain them within [0,1] as well. The resulting distributions are shown in Figure~\ref{fig:intensity_plot}.

\subsubsection{Effect on our Attacks}
Our adversarial augmentations perturb also the intensity signals. As we limited the maximum shift of intensity to 0.3, in theory this allowed to transition from the mean \textit{car} value (lowest for SemanticKITTI), to the mean \textit{traffic-sign} value (highest). However, 0.3 prevented from fully representing the the intensity distribution of \textit{traffic-sign} starting from \textit{car}. Therefore, while an untargeted attack on \textit{car} could focus on weaker boundaries (both semantically and in terms of intensity values), such as \textit{truck} and \textit{other-vehicle}, not all targeted attacks can be very effective due to the intensity shift. For example, \textit{car} and \textit{traffic-sign} are rather far (both semantically and in terms of intensity values), rendering such targeted attack more difficult to achieve. For these reasons, we focused our targeted attacks on traffic participants other than \textit{car}, such as \textit{other-vehicle} and \textit{bicycle}.

\begin{table*}[t!]
\begin{center}
\begin{tabular}{ll|ll|cccc}
\toprule

LiDAR Intensity Transformation & Dataset & ID & Method & mIoU & \textit{car} & \textit{pers.} & \textit{bike} \\

\midrule

\multirow{4.4}{*}{none} & \multirow{2}{*}{SemanticKITTI} & c.o & Cylinder3D & \textbf{64.39} & 96.02 & 73.84 & \textbf{47.15} \\
&& u.c & [ours] &  63.85 & \textbf{96.03} & \textbf{77.11} & 33.41 \\
\cmidrule(lr){2-8}
& \multirow{2}{*}{$\rightarrow$ Waymo} & c.o & Cylinder3D & 29.43 & 72.10 & 6.69 & \textbf{5.68} \\
&& u.c & [ours] &  \textbf{37.74} & \textbf{82.38} & \textbf{17.13} & 5.04 \\

\cmidrule(lr){1-8}

\multirow{4.4}{*}{all 0} & \multirow{2}{*}{SemanticKITTI} & c.o & Cylinder3D & 12.27 & 23.15 & 0.61 & 0.01 \\
&& u.c & [ours] &  \textbf{31.09} & \textbf{83.13} & \textbf{6.77} & \textbf{0.25} \\
\cmidrule(lr){2-8}
& \multirow{2}{*}{$\rightarrow$ Waymo} & c.o & Cylinder3D & 22.75 & 53.19 & 0.05 & \textbf{0.15} \\
&& u.c & [ours] &  \textbf{32.43} & \textbf{75.43} & \textbf{1.49} & 0.01 \\

\cmidrule(lr){1-8}

\multirow{4.4}{*}{Gaussian noise (std: 0.3)} & \multirow{2}{*}{SemanticKITTI} & c.o & Cylinder3D & 15.59 & 61.91 & 26.84 & 1.61\\
&& u.c & [ours] &  \textbf{32.51} & \textbf{77.98} & \textbf{35.06} & \textbf{11.50}\\
\cmidrule(lr){2-8}
& \multirow{2}{*}{$\rightarrow$ Waymo} & c.o & Cylinder3D & 21.64 & 62.18 & 13.66 & 6.16\\
&& u.c & [ours] &  \textbf{34.22} & \textbf{70.01} & \textbf{35.77} & \textbf{6.77}\\

\cmidrule(lr){1-8}

\multirow{4.4}{*}{uniform random $[0, 1]$} & \multirow{2}{*}{SemanticKITTI} & c.o & Cylinder3D & 8.73 & 11.33 & \textbf{20.53} & 2.24 \\
&& u.c & [ours] &  \textbf{21.61} & \textbf{69.98} & 8.89 & \textbf{7.91} \\
\cmidrule(lr){2-8}
& \multirow{2}{*}{$\rightarrow$ Waymo} & c.o & Cylinder3D & 5.76 & 0.91 & 0.31 & 0.59 \\
&& u.c & [ours] &  \textbf{23.10} & \textbf{68.82} & \textbf{12.50} & \textbf{1.95} \\

\cmidrule(lr){1-8}

\multirow{4.4}{*}{uniform random noise $+[0,+0.3]$} & \multirow{2}{*}{SemanticKITTI} & c.o & Cylinder3D & 28.21 & 89.40 & 35.98 & \textbf{24.30} \\
&& u.c & [ours] &  \textbf{44.88} & \textbf{94.76} & \textbf{49.16} & 23.52 \\
\cmidrule(lr){2-8}
& \multirow{2}{*}{$\rightarrow$ Waymo} & c.o & Cylinder3D & 35.36 & 65.39 & 37.75 & 12.29\\
&& u.c & [ours] &  \textbf{42.31} & \textbf{80.64} & \textbf{48.48} & \textbf{12.36}\\

\cmidrule(lr){1-8}

\multirow{4.4}{*}{uniform random noise $+[-0.3,+0.3]$} & \multirow{2}{*}{SemanticKITTI} & c.o & Cylinder3D & 41.94 & 92.89 & 35.30 & 10.21 \\
&& u.c & [ours] & \textbf{54.21} & \textbf{95.09} & \textbf{68.93} & \textbf{27.63} \\
\cmidrule(lr){2-8}
& \multirow{2}{*}{$\rightarrow$ Waymo} & c.o & Cylinder3D & 32.11 & 70.45 & 19.85 & 8.13 \\
&& u.c & [ours] & \textbf{41.02} & \textbf{81.09} & \textbf{36.06} & \textbf{8.36} \\

\cmidrule(lr){1-8}

\multirow{4.4}{*}{random shift $\pm 0.3$} & \multirow{2}{*}{SemanticKITTI} & c.o & Cylinder3D & 17.54 & 71.99 & 27.36 & 1.62 \\
&& u.c & [ours] & \textbf{35.93} & \textbf{81.06} & \textbf{40.58} & \textbf{10.95} \\
\cmidrule(lr){2-8}
& \multirow{2}{*}{$\rightarrow$ Waymo} & c.o & Cylinder3D & 27.05 & 61.96 & 33.64 & \textbf{8.12} \\
&& u.c & [ours] & \textbf{38.71} & \textbf{76.21} & \textbf{46.83} & 6.92 \\



\bottomrule

\end{tabular}
\caption{Detailed impact of the LiDAR intensity as extra input on \protect\culine{3D semantic segmentation}. Various transformations are applied to the intensity values (first column), leading to the reported IoUs. All models are trained on SemanticKITTI and based on the Cylinder3D architecture~\cite{zhu2021cylindrical}, with [ours] trained with our untargeted adversarial augmentations only on \textit{car} points, also for the intensity values (restricted to a maximum perturbation of 0.3). Across the various datasets and input configurations, a total of only two models is evaluated in this table: one for the standard Cylinder3D (c.o), and one trained with our adversarial augmentation method (u.c).}
\label{table:intensity_sensitivity}
\end{center}
\end{table*}

\subsection{Robustness to Changing LiDAR Intensity Values}\label{sec:results_intensity_robustness}
In Table~\ref{table:intensity_sensitivity} we assess the increase of robustness by our adversarial augmentations when changing the LiDAR intensity values, while leaving the 3D points unchanged. We do so by comparing our method with Cylinder3D and applying various transformations to the intensity values in input. We report such results both in-domain (i.e., SemanticKITTI) and out-of-domain (i.e., Waymo). The models evaluated in the table were trained only on SemanticKITTI, and our adversarial augmentations were applied only on \textit{car} points in an untargeted fashion, leaving all the other points unchanged compared to the baseline, which also applied standard augmentations (e.g., rotation).

\subsubsection{Impact of Using the LiDAR Intensity}
As discussed, the intensity is a powerful input signal, which can help distinguishing the semantic classes. This is why models trained with it are particularly sensitive to changes of its values, which caused the transfer on Waymo to be more effective when training without intensity (Table~\ref{table:intensity_sensitivity}). The table shows how much Cylinder3D relies on the reflectivity value, when trained both with and without our adversarial augmentations. Our method outperforms the standard Cylinder3D across all changes applied, proving the added robustness of our augmentations. The gap is most extreme in the uniform random case ranging from 0 to 1 (all possible intensity values): on SemanticKITTI our method achieved a 6.2x higher IoU on \textit{car} (70.0 compared to 11.3) and 2.5x higher mIoU (21.6 - 8.7); on Waymo ours reached a 76x higher IoU on \textit{car} (68.8 - 0.9) and 4x higher mIoU (23.1 - 5.8). However, it is the complete lack of intensity values (all 0) that shows how much each model relies on the intensity signal after training with it. In this case, for the reference class \textit{car}, the IoU of our approach dropped only by 13 for SemanticKITTI and 7 for Waymo, compared to the IoU of Cylinder3D dropping by 73 and 19 respectively.

\subsubsection{Benefits of our Method}
Similarly to the improved robustness on different vehicle shapes thanks to geometrical perturbations (e.g., rare and damaged, with CrashD and Waymo for 3D object detection), the reason for the significant improvements shown in Table~\ref{table:intensity_sensitivity} is that, when augmenting, our method alters also the intensity values in an adversarial fashion. Our augmented model learns to be less reliant on the reflectivity input, allowing it to generalize better to different values (e.g., uniform random). On the reference class \textit{car}, despite the extreme intensity changes applied in input, our model always reached in- or out-of-domain a remarkable IoU of at least 68.82 (lowest reached with uniform random [0,1], on Waymo). Conversely, Cylinder3D trained without our adversarial augmentations was unable to deliver satisfactory predictions, with the \textit{car} IoU dropping as low as 0.91 (lowest reached with uniform random [0,1], on Waymo).

The benefits of our approach are evident also considering the in-domain performance alone (Table~\ref{table:intensity_sensitivity}). On the standard SemanticKITTI (none), the baseline achieved a higher mIoU than our method. However, any transformation applied to the intensity values introduced major changes, with ours always outperforming Cylinder3D by significant margins.

\subsubsection{Side-effect on Other Classes}
Despite our adversarial modifications were only applied on \textit{car} points in an untargeted way, as shown in Table~\ref{table:intensity_sensitivity} they had a major effect on other classes as well, such as for \textit{person} and \textit{bicycle}. For \textit{person}, our method improved the IoU by up to 40x on Waymo (reached with uniform random [0,1]) and by up to 11x on SemanticKITTI (reached with all 0). Analogously, for \textit{bicycle}, despite performing worse than Cylinder3D both in- and out-of-domain without modifying the intensity values in input (none), our approach improved the IoU by up to 7x on SemanticKITTI (reached with Gaussian) and 3.3x on Waymo (reached with uniform random [0,1]) when altering the intensity values.

\subsubsection{In-domain and Out-of-domain}
Furthermore, while the mIoU gap between in- and out-of-domain is relatively large without modifying the intensity values, this gap shrinks and even inverts when applying many of the modifications shown in Table~\ref{table:intensity_sensitivity}. With Gaussian noise, both Cylinder3D and our approach achieved a higher mIoU on Waymo than SemanticKITTI, albeit on a different number of classes. Apart from the constrained uniform random noises, all other modifications made our method achieve a higher mIoU on Waymo than SemanticKITTI. This can be attributed to the denser point cloud in Waymo compared to SemanticKITTI, which allows to extract better geometrical features when LiDAR intensity values become unreliable due to the changes.

\subsubsection{Reducing the Domain Gap}
Interestingly, in Table~\ref{table:intensity_sensitivity} there are random modifications that significantly improved the transfers to Waymo in terms of mIoU (up to +6, compared to the unmodified intensity inputs). This was the case of the constrained uniform random noise settings ($+[0,+0.3]$ and $+[-0.3,+0.3]$), for both Cylinder3D and our model. This improvement can be attributed to the transformation bringing the intensity distribution of Waymo closer to that of the standard SemanticKITTI. Therefore, the random transformation reduces the intensity gap shown in Figure~\ref{fig:intensity_plot} between the two datasets, as Waymo typically displayed a lower intensity compared to SemanticKITTI. For the same reasons, ours benefitted also from the random shift $\pm 0.3$ in terms of mIoU.


\begin{figure*}[t]
\centering
  \includegraphics[width=1.0\textwidth]{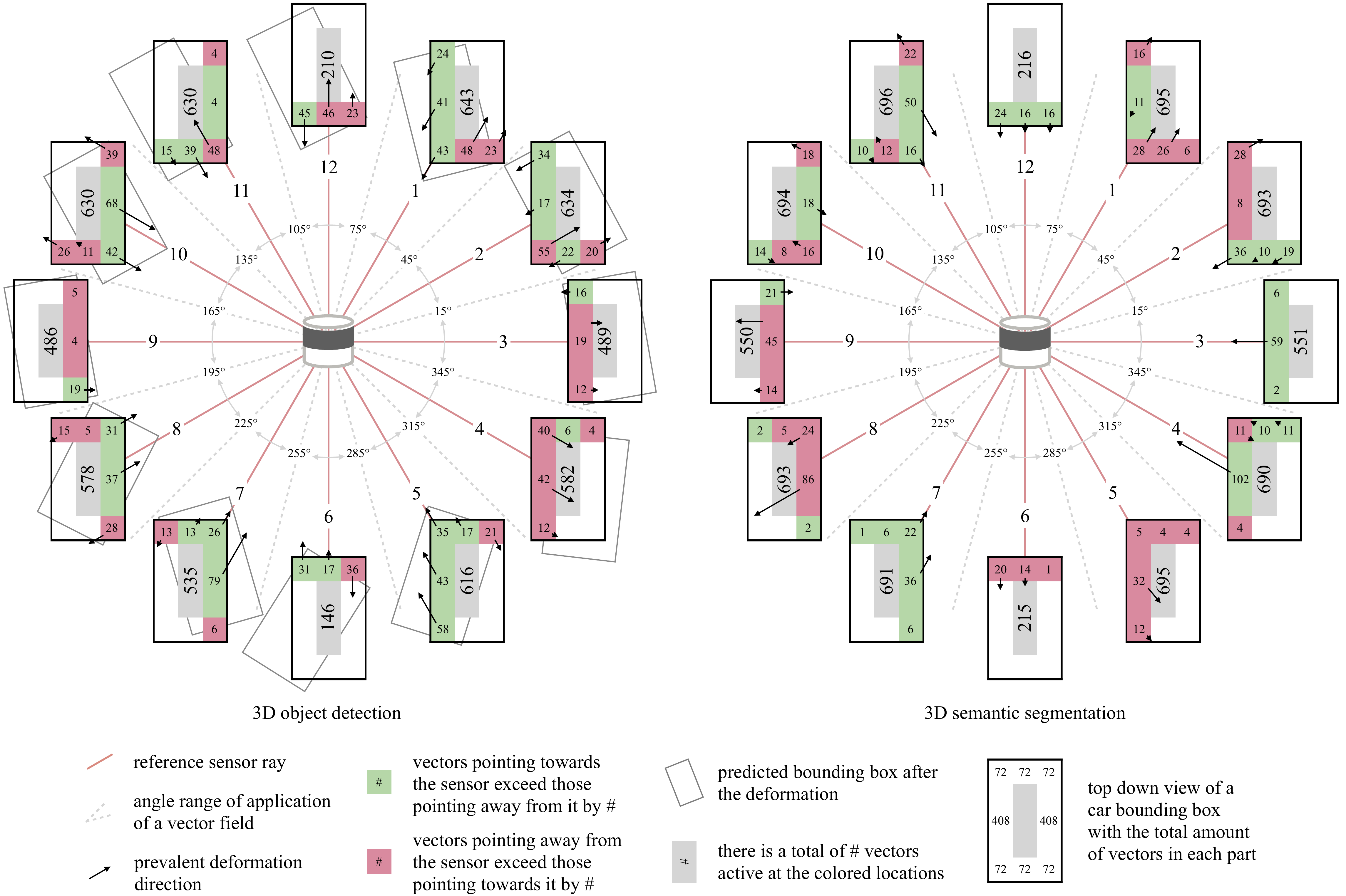}
   \caption{Comparison of adversarial vector fields learned for 3D object detection (left) and 3D semantic segmentation (right).
   For each task, $G=12$ vector fields are displayed according to their relative rotation from the sensor. All are trained to reduce the performance on the \textit{car} class.
   The color in each area indicates the prevalent direction of deformation: green means that the majority of vectors points towards the sensor, while red means away from it. This is emphasized by the black arrows, which are aligned to the sensor rays and have a magnitude proportionate to the number inside the corresponding colored areas. This number indicates how many more vectors point in that direction compared to those facing the opposite one, after being projected along the sensor view ray (those with magnitudes lower than their random initialization are ignored).
   Object detection vectors often aim at rotating the predicted bounding box, which is represented in grey as an estimation, thereby reducing the AP. They achieve this by pushing one corner of the car and pulling the opposite one.
   Instead, segmentation vectors do not exhibit this behavior, because rotating the objects does not reduce the semantic IoU. In comparison, they need to alter more points, which makes them significantly more active, as shown by the number in the middle of each vector field.
   }
   \label{fig:vectors_comp}
\end{figure*}

\subsection{Cross-task Vectors Comparison}\label{sec:results_vectors_comp}
Figure~\ref{fig:vectors_comp} shows an analysis of the vector fields learned for each of the two tasks. In this section we discuss the effects of the tasks on the vector fields.

\subsubsection{Object Detection}
For 3D object detection, the adversarial loss aims at reducing the confidence and the IoU of the predicted box with respect to the ground truth box. The vectors can achieve this by simply making the detector believe that the object is rotated. Specifically, as shown in Figure~\ref{fig:vectors_comp}, the vectors learned to pull a corner of the car towards the sensor (green) and push the opposite one away (red). This changes the orientation of the surface on which the detector bases its regression of the rotation. Therefore, the detector predicts a rotated box, which significantly reduces the IoU. This affects the AP as the amount of boxes exceeding the IoU threshold decreases. This behavior is visible across most of the $G=12$ relative positions, particularly evident at 6 and 12. All other positions also exhibit the same actions, apart from 4. At 4, all corners push the points away. Nevertheless, this also reduces the box IoU as it shifts all points in the same direction.

\setlength{\tabcolsep}{8.2pt}
\begin{sidewaystable}
\begin{center}
\begin{minipage}{\textheight}
\begin{tabular*}{\textheight}{ll|cc|cccc|cccc|cccc}
\toprule

& \multicolumn{1}{c|}{~} & \multicolumn{2}{c|}{SemK attack} & \multicolumn{4}{c|}{SemanticKITTI} & \multicolumn{4}{c|}{$\rightarrow$ Waymo} 
& \multicolumn{4}{c}{$\rightarrow$ nuScenes} 
\\

ID & \multicolumn{1}{l|}{Method} & mIoU & \textit{car}  & mIoU & \textit{car} & \textit{pers.} & \textit{bike} & mIoU & \textit{car} & \textit{pers.} & \textit{bike} 
& mIoU & \textit{car} & \textit{pers.} & \textit{bike}
\\

\midrule

c.n & Cylinder3D & -&-& 59.23 & \textbf{96.23} & 68.67 & \textbf{32.61} & 
37.39 & 63.74 & 37.95 & 4.05 &
32.11 & 56.17 & 0.04 & 0.35
\\

u.d & [ours] for obj.det. & 58.54 & 94.47 &  58.43 & 95.73 & \textbf{70.90} & 17.19 & 36.28 & \textbf{70.76} & 30.71 & 5.99 & 
32.27 & 62.49 & 0.09 & 1.83 \\

u.n & [ours] for sem.seg. & \textbf{56.63} & \textbf{86.03} & \textbf{59.40} & 95.73 & 70.35 & 26.53 & 
\textbf{40.36} & 70.70 & \textbf{52.73} & \textbf{7.57} &
\textbf{34.30} & \textbf{63.13} & \textbf{1.02} & \textbf{2.52}
\\

\bottomrule

\end{tabular*}
\caption{Comparison on \protect\culine{3D semantic segmentation} of adversarial vectors learned for 3D object detection and 3D semantic segmentation. All models are based on Cylinder3D~\cite{zhu2021cylindrical} and do not use the LiDAR intensity.
}
\label{table:cvpr_comp_on_semantic}
\end{minipage}
\end{center}
\end{sidewaystable}
\setlength{\tabcolsep}{16pt}
\begin{sidewaystable}
\begin{center}
\begin{minipage}{\textheight}
\begin{tabular*}{\textheight}{ll|ccc|c|c|cc|cc}
\toprule

& \multicolumn{1}{c|}{~} & \multicolumn{4}{c|}{KITTI} & $\rightarrow$ Waymo & \multicolumn{4}{c}{$\rightarrow$ CrashD} \\

& \multicolumn{1}{c|}{~} & \multicolumn{3}{c|}{AP} &  & & \multicolumn{2}{c|}{AP \textit{normal}} & \multicolumn{2}{c}{AP \textit{rare}} \\
ID & \multicolumn{1}{l|}{Method} & \textit{easy} & \textit{mod.} & \textit{hard} & ASR & AP & \textit{clean} & \textit{crash} & \textit{clean} & \textit{crash} \\

\midrule

p.p & PointPillars &  \textbf{88.24} & 77.11 & 74.55 & - & 40.86 & 65.20 & 43.67 & 34.14 & 22.48  \\

p.e & [ours] for sem.seg. & 86.88	& 75.98	& 68.73	& 17.2 & 40.70 & 59.20 & 42.82 & 39.23 & 28.15 \\

p.o & [ours] for obj.det. & 87.05 & \textbf{77.13} & \textbf{75.55} & \textbf{63.4} & \textbf{44.61} & \textbf{67.95} & \textbf{52.87} & \textbf{43.40} & \textbf{30.37} \\

\bottomrule

\end{tabular*}
\caption{Comparison on \protect\culine{3D object detection} of adversarial vectors learned for 3D object detection (\cite{lehner20223dvfield}) and 3D semantic segmentation. All models are based on PointPillars~\cite{lang_pointpillars_2019}.}
\label{table:cvpr_comp_on_det}
\end{minipage}
\end{center}
\end{sidewaystable}

\subsubsection{Semantic Segmentation}
Interestingly, this rotatory pattern is mostly absent on the vector fields learned for 3D semantic segmentation, which exhibit a rather different behavior (Figure~\ref{fig:vectors_comp}). This can be explained by considering the two different tasks, adversarial losses and metrics involved.
For object detection, the metric is based on the IoU between the predicted box and the corresponding ground truth. AP is relatively discrete, as it is computed by counting positive and negative predictions. Moreover, an object is considered missed if the IoU is lower than a certain threshold (e.g., 0.7).

\subsubsection{AP vs.~IoU}
While pretending to rotate the object can be detrimental for the AP and lead to a successful adversarial attack for object detection, it is not as useful for semantic segmentation. Semantic segmentation is a dense task, demanding a prediction for each point. Every prediction and every point count towards the IoU, which makes it rather smooth compared to the AP. Therefore, to change the predicted class of a 3D point, the vectors here cannot simply rotate the object, but need to make it look as if it belonged to a different class. This makes it significantly harder for semantic segmentation compared to object detection.

\subsubsection{Number of Activated Vectors}
Semantic segmentation activated more vectors than object detection, because it required to move more points to fool the model and reduce the IoU. This is shown in Figure~\ref{fig:vectors_comp} through the values in the middle of each position. For semantic segmentation, a maximum of only 6 vectors within a vector field remained inactive (reached at position 4: 690 out of 696), for a total of 7079 active vectors out of 7104. Conversely, for object detection up to 118 vectors remained inactive (reached at position 7: 578 out of 696), for a total of 6179 active out of 7104. Therefore, semantic segmentation activated 900 more vectors, equivalent to 11\% for the 12 positions, or a relative 15\% increase over object detection. For this analysis, as highlighted in the figure by the colors, we considered visible vectors from the sensor, and we regarded as inactive those having a magnitude lower than that of their random initialization.

\subsubsection{Impact of $G$ on the Vectors}
Figure~\ref{fig:vectors_comp} visually explains also the reasons why with $G=1$, i.e., using 1 vector field for the entire dataset, the vectors can reach only sub-optimal results (Table~\ref{table:relative_rotations}). Although with $G=1$ the vectors can still learn to give the illusion of a rotated object (i.e., having all corners alternating between pulling and pushing the points), they cannot achieve the diversity and specificity shown in the figure with $G=12$, and they cannot be as directed as those with $G=12$. By being effective in all positions and distances, the vector field with $G=1$ needs to contain vectors pointing in all directions. This limits their efficacy as their magnitude decreases when projected to the sensor view rays. Therefore, the shifts and rotations applied with $G=1$ cannot be as strong as those with $G=12$ at all positions. 

\subsubsection{Transferring the Attack to a Different Task}
The difference between the vector fields learned for the two tasks is shown also in Tables~\ref{table:cvpr_comp_on_semantic} and~\ref{table:cvpr_comp_on_det}. Interestingly, augmenting with the vectors learned for object detection (\cite{lehner20223dvfield}) when training a semantic segmentation model delivered strong predictions for the reference class \textit{car}, performing on par or even better than the semantic vectors, also when transferring to Waymo (Table~\ref{table:cvpr_comp_on_semantic}). Although the detection vectors lowered the mIoU, this shows the flexibility of our approach even across tasks. Augmenting PointPillars with the vectors learned for segmentation underperformed the baseline on \textit{car}, while they significantly improved the AP on the challenging \textit{rare} samples of CrashD (Table~\ref{table:cvpr_comp_on_det}).

This difference between the vectors learned for the two tasks could be attributed to the application of the detection vectors resulting in pseudo rotated cars, instead of the semantic vectors trying to resemble a different class. Augmenting with the detection vectors, thanks to the pseudo rotatory behavior shown in Figure~\ref{fig:vectors_comp}, is likely to improve the regression of the bounding box rotation. Since estimating the rotation correctly is crucial to exceed the IoU threshold with the ground truth box, improving the regression of the rotation translates in better box IoU on challenging cases (e.g., AP on Waymo in Table~\ref{table:cvpr_comp_on_det}). Instead, by not exhibiting that rotatory behavior, the semantic vectors are likely to not improve the rotation regression (Table~\ref{table:cvpr_comp_on_det}). On the proposed CrashD, the semantic vectors underperformed the detection baseline on \textit{normal} cars, but significantly outperformed it on difficult \textit{rare} ones, resembling long tail samples, both damaged and undamaged.

\begin{table*}[t]
\begin{center}
\begin{tabular}{lll|cc|cc}
\toprule

& \multicolumn{2}{c|}{~} & \multicolumn{2}{c|}{$\rightarrow$ Waymo (2 cl.)} & \multicolumn{2}{c}{$\rightarrow$ nuScenes (10 cl.)} \\

ID & Method & DA & mIoU & rel.change & mIoU & rel.change \\

\midrule

c.l.b & Complete\&Label baseline & no & 46.3 & - & 27.9 & -\\
c.l & Complete\&Label & yes & 52.0 & +12.3\% & \textbf{31.6} & \textbf{+13.3\%} \\
\cmidrule(lr){1-7}
c.n & Cylinder3D & no & 51.8 & - & 25.8 & - \\
u.n & [ours] & \textbf{no} & \textbf{63.4} & \textbf{+22.4\%} & 28.0 & +8.5\% \\

\bottomrule
\end{tabular}
\end{center}
\caption{Comparison with domain adaptation (DA) for \protect\culine{3D semantic segmentation}. Models transferred to Waymo and nuScenes after training without LiDAR intensity on SemanticKITTI with 10 classes (Complete\&Label~\cite{yi2021complete_and_label}) and 19 classes (ours). The evaluation follows the class setup of~\cite{yi2021complete_and_label}. Due to the major differences between Complete\&Label and our approach, also in terms of baselines, we report the relative improvements (rel.change).}
\label{table:comp_adaptation_semantic}
\end{table*}

\textbf{Detection vectors for segmentation}
As described in Section~\ref{sec:specificity-generalization}, for our adversarial augmentation method the ability to generalize is related to the performance of the adversarial attacks on which the augmentation is based upon. The detection vectors on the semantic model delivered poor attack performance (SemK attack), with only a slight decrease in IoU (Table~\ref{table:cvpr_comp_on_semantic}). This means that the examples generated by perturbing \textit{car} points with the detection vectors were not challenging for the semantic model (i.e., Cylinder3D). Although this would hint towards the futility of these vectors in semantic settings, using them as augmentation delivers significant improvements on \textit{car} segments on both Waymo and nuScenes. Therefore, despite being weak as attacks, the detection vectors successfully regularized the training of Cylinder3D by augmenting the training data beyond standard augmentations.

\textbf{Segmentation vectors for detection}
The semantic vectors delivered a stronger cross-task attack, with an ASR of 17.2 against PointPillars (Table~\ref{table:cvpr_comp_on_det}). This can be attributed to the semantic ones trying to shift the points to resemble a different class, instead of rotating the object. As the object detector has no explicit semantic understanding beyond its reference class \textit{car}, augmenting with the semantic vectors probably expanded the training data in a direction which is not beneficial for the detector, causing confusion.
Nevertheless, optimizing the adversarial vectors for each task independently leads to more significant deformations targeted around the weaknesses of the respective model and task combination, allowing for more meaningful adversarial augmentations, which lead to better overall performance (mIoU and AP).

\subsubsection{Considerations on adversarial augmentation}
The essence of data augmentation is enriching the training data. Whether adversarial (e.g., semantic vectors in Table~\ref{table:cvpr_comp_on_semantic}), or not (e.g., detection vectors, which were learned for a different task and model, so they lost their adversarial properties), data augmentation is crucial for domain generalization, to tackle challenging out-of-domain samples. In particular, we believe that data augmentation is most effective when it manages to push the boundaries of the training distribution towards corner cases which are not present in the original training set.

While this expansion could be easily achieved via extreme transformations of the input, it is not necessarily beneficial for the model and may even harm its performance, as seen for the sample-specific adversarial approaches (Section~\ref{table:main_semantic}). The difficulty is preserving the usefulness of the augmented samples towards the task at hand and real-world data. This is where our plausibility constraints (Section~\ref{sec:constraints}) play a fundamental role. Together with our generated examples, which represent hard cases by being adversarial, the constraints determine a good balance between expanding the training data and preserving plausibility of the new samples. When using the vectors across different tasks, the plausibility is still preserved as all constraints are applied, but the generated examples do not represent hard cases as the vectors were optimized for a different task. This ultimately reduces the degree of expansion of the training distribution given by the augmentations, compared to using purposed vectors optimized for the same task.

\begin{figure*}[t]
\centering
  \includegraphics[width=1.0\textwidth]{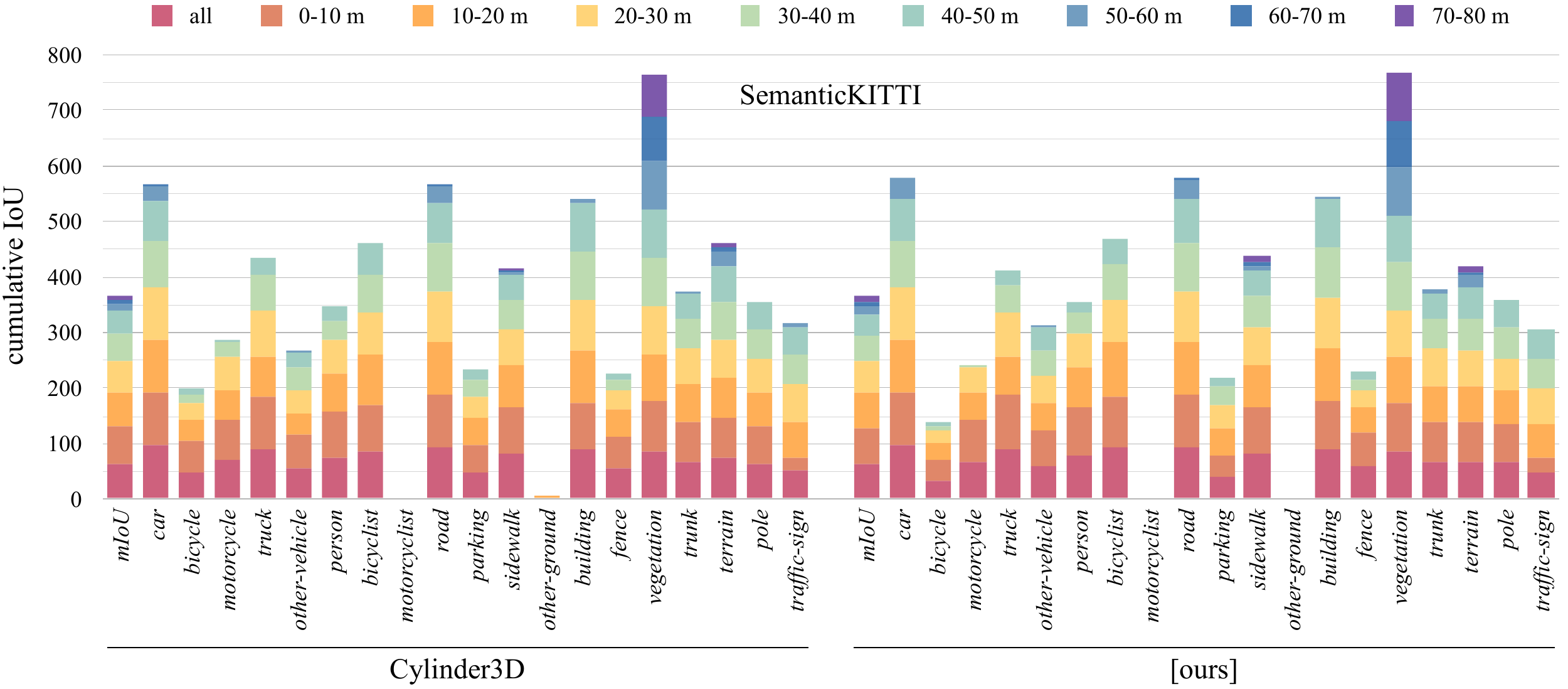}
   \caption{Evaluation by distance on the validation set of SemanticKITTI~\cite{behley2019semantickitti}. Predictions of Cylinder3D~\cite{zhu2021cylindrical} trained without (left, with ID c.0) and with (right, with ID u.c) our adversarial augmentations. Both models use the intensity signal. Our approach augments only \textit{car} points in an untargeted fashion. The plot shows stacked bars  (i.e., cumulative IoU) and the colors represent different distance bins, as indicated in the legend above.}
   \label{fig:distance_plot_SK}
\end{figure*}

\begin{figure*}[t]
\centering
  \includegraphics[width=1.0\textwidth]{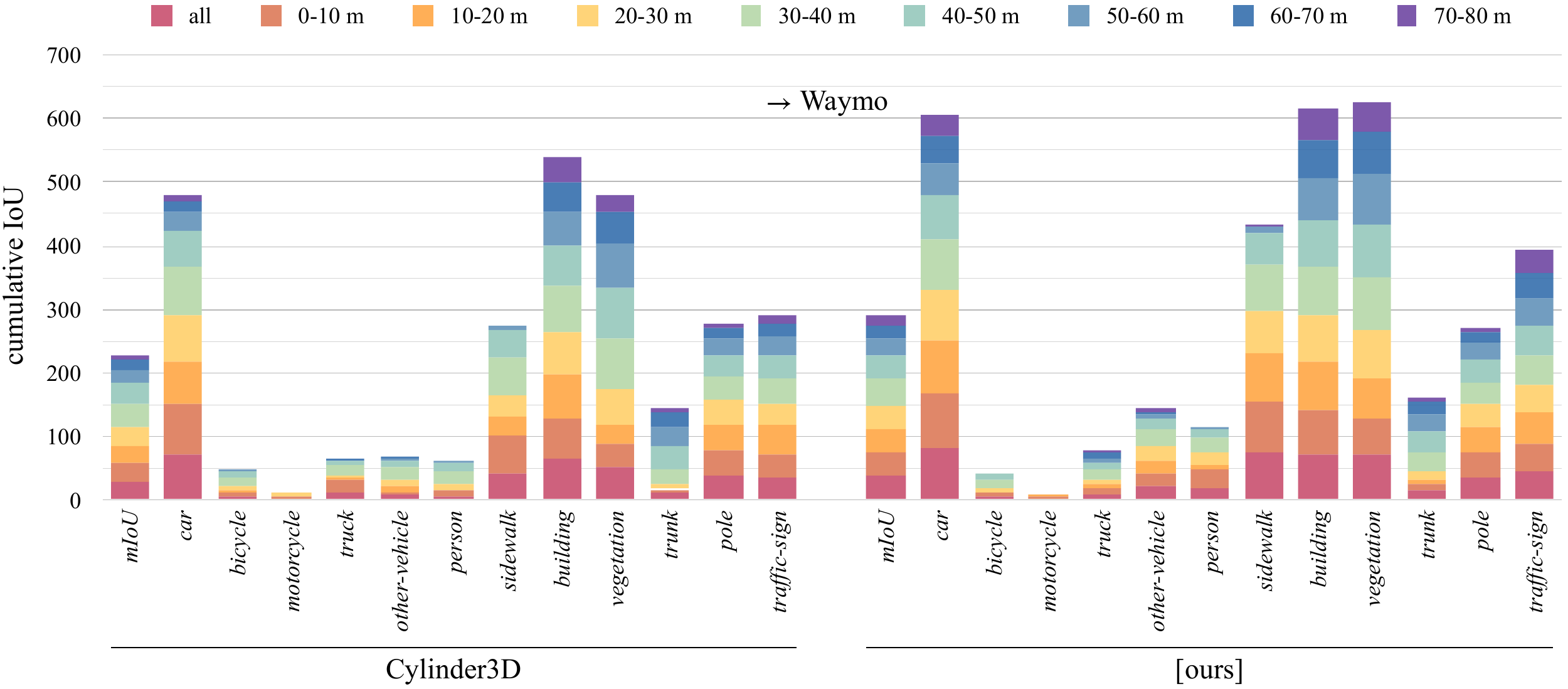}
   \caption{Evaluation by distance on the transfer to the validation set of Waymo~\cite{sun2020waymo}. Predictions of Cylinder3D~\cite{zhu2021cylindrical} trained on SemanticKITTI without (left, with ID c.0) and with (right, with ID u.c) our adversarial augmentations. Both models use the intensity signal. Our approach augments only \textit{car} points in an untargeted fashion. The plot shows stacked bars (i.e., cumulative IoU) and the colors represent different distance bins, as indicated in the legend above.}
   \label{fig:distance_plot_W}
\end{figure*}

\subsection{Comparison with Domain Adaptation}\label{sec:results_DA_comp}
Our approach does not use any information about the target data (e.g., Waymo and nuScenes), making it a domain generalization method. Conversely, domain adaptation bridges the domain gap by exploiting knowledge about the target data, such as the average object size (\cite{wang2020train}).

\textbf{Semantic segmentation}
In Table~\ref{table:comp_adaptation_semantic} we show a comparison with the domain adaptation work for LiDAR point clouds of \cite{yi2021complete_and_label}, who exploited the different sensors used to captured the data. Compared to all other experiments, where we reported on all matching classes, in these experiments we use the setup of~\cite{yi2021complete_and_label}, who used 2 classes for Waymo, namely \textit{person} and the super-class \textit{vehicle}, and 10 for nuScenes. However, while we trained our models on all 19 classes of SemanticKITTI, \cite{yi2021complete_and_label} trained only on the 10 classes they considered for nuScenes, making it easier for them to transfer their models.
Due to the performance discrepancy of their baseline compared to ours (i.e., Cylinder3D), we consider the relative improvements of each approach.

Despite not using any target information, our adversarial augmentation approach outperformed the domain adaptation method of~\cite{yi2021complete_and_label} by a significant margin on Waymo. Ours improved by 22\% over our baseline (i.e., Cylinder3D), while theirs improved by 12\% over their baseline. Both baselines were trained on SemanticKITTI and transferred without any fine-tuning nor domain adaptations. However, their models were trained only on the 10 classes being evaluated, compared to ours trained on the full set of 19 classes, which made it more challenging for ours.
Transferring to nuScenes, due to the sparsity of its point clouds compared to the source data (i.e., SemanticKITTI), Cylinder3D was unable to deliver satisfactory results, underperforming the baseline of~\cite{yi2021complete_and_label}. Over the baselines, ours improved by 8.5\%, compared to their 13.3\% increase. Therefore, using knowledge about the target domain (i.e., nuScenes) was helpful here.

Moreover, our method does not alter the entire point clouds, but only a single object per scene, always belonging to the same class (i.e., \textit{car} in this case). Instead, domain adaptation approaches, such as that of~\cite{yi2021complete_and_label} act upon the whole 3D point clouds. This allows them to improve over all classes, while ours mainly focuses on a single class (e.g., \textit{car}). This could be an extra reason why ours outperformed on the 2 class setting (i.e., Waymo), but not with 10 classes (i.e., nuScenes).
Nevertheless, on Waymo, without using any knowledge about the target data nor sensor, ours outperformed a recent domain adaptation technique designed to be robust across different LiDAR sensors~\cite{yi2021complete_and_label}. This confirms the benefits of our approach towards robustness and out-of-domain data.

\subsection{Combination with Domain Adaptation}\label{sec:results_DA_combination}
By addressing domain generalization, our approach does not use any target information. Therefore, ours is not alternative to domain adaptation methods~\cite{wang2020train, yi2021complete_and_label}, which make use of target data. However, similarly to other data augmentation strategies, our approach can be combined with domain adaptation techniques.

\textbf{Object detection}
As shown in Table~\ref{table:AP_on_all}, such combination further boosts the performance on challenging out-of-domain data. By altering the objects size via the statistical normalization (SN) of~\cite{wang2020train}, the AP on Waymo increased. Constrained by the high amount of false positives and negatives, when combined with SN, ours retained a margin of over 2\% compared to PointPillars with SN.
Moreover, the AP on CrashD improved significantly across all categories, especially for the hardest \textit{rare crash} group. The results show how, despite a substantial increase in AP from PointPillars~\cite{lang_pointpillars_2019}, SN alone did not reach the full potential of the detector. Only when combined with ours, the AP doubled (\textit{normal crash}) and more than tripled (\textit{rare crash}) over PointPillars, without using any extra information about the target. This shows the benefit of this combination, and reiterates the added value of incorporating adversarially deformed objects via data augmentation to improve generalization to out-of-domain samples.




\subsection{Different 3D Object Detectors}\label{sec:results_3D_detectors}
In Table~\ref{table:AP_on_all}, we also compare the performance of our augmentations when paired with different 3D object detectors, namely PointPillars~\cite{lang_pointpillars_2019} and Part-A$^2$~\cite{shi2019parta2}. Remarkably, using the proposed adversarial augmentation improved the AP of Part-A$^2$ on Waymo by a large margin. The superiority of Part-A$^2$ over the other detector can be attributed to its part-awareness~\cite{shi2019parta2}, which might have set its focus on the most relevant object parts (e.g., wheels) and their relationships to identify cars also in out-of-domain settings.
Adding our adversarial deformations significantly improved the generalization of both detectors to out-of-domain data, despite training our vector fields solely against PointPillars, and transferring them to Part-A$^2$. This shows the wide applicability and transferrability of our techniques.

\subsection{Robustness at Further Distances}\label{sec:results_distances}
\textbf{Semantic segmentation} In Figures~\ref{fig:distance_plot_SK} and~\ref{fig:distance_plot_W} we compare the IoUs obtained by Cylinder3D with those from our method at varying distances from the LiDAR sensor. Figure~\ref{fig:distance_plot_SK} reports the results on SemanticKITTI, while Figure~\ref{fig:distance_plot_W} those of the transfer to Waymo. Towards this end, we first computed the distance in 3D of each point to the LiDAR sensor, we clustered them in 8 bins ranging 10 meters each, and finally computed the IoUs separately for each bin.

Figure~\ref{fig:distance_plot_SK} shows that our method performed similarly to the baseline (i.e., Cylinder3D) on SemanticKITTI across various distances, with only minor differences. This is reflected by the similar mIoUs in Table~\ref{table:main_semantic}.
However, Figure~\ref{fig:distance_plot_W} shows a different outcome for the challenging out-of-domain Waymo. Despite reaching similar IoUs at closer distances, our adversarial augmentations delivered better predictions, especially further from the sensor, i.e., more challenging sparser areas, where the LiDAR intensity follows an opposite trend to that of SemanticKITTI (Section~\ref{sec:results_intensity_distributions}). This superiority is shown by the amount of blue and purple in the plot, which correspond to the furthest bins, between 60 and 70 meters, and between 70 and 80, respectively. Especially for \textit{car}, \textit{building}, and \textit{vegetation}, our method managed to deliver good quality predictions, even beyond 70 meters. Instead, the standard Cylinder3D between 70 and 80 meters could only recognize a few \textit{car} points, with an IoU of 6.7, compared to 32.9 of our approach.
This confirms the robustness of our method, also at high distances.

\subsection{Combination with Data Augmentations}\label{sec:results_combination_dataaug}
Being an adversarial data augmentation, our approach is not alternative to other augmentation techniques, but can be applied in combinations with others, e.g., Mix3D~\cite{nekrasov2021mix3d}.

\textbf{Object detection} As shown in Table~\ref{table:AP_on_all}, on KITTI removing all augmentations had a major impact on the AP. For PointPillars~\cite{lang_pointpillars_2019}, not using augmentations (no augm.) drastically reduced the APs, especially on CrashD at IoU 0.7. Instead, at IoU 0.5, the AP on \textit{normal clean} was 65.59, with the standard PointPillars reaching 98.91. Adding common augmentations (no obj.~sampl., e.g., flip and rotation) increased the APs, but introducing the popular object sampling~\cite{lang_pointpillars_2019} (PointPillars) improved them even further.
Nevertheless, applying our adversarial augmentations on \textit{car} points on top of the standard augmentations significantly improved the IoUs on out-of-domain data. This shows the compatibility of our adversarial approach with other data augmentation techniques.

\subsection{Impact of Data Annotations}
\label{sec:exp_annotations}
In Table~\ref{table:annotations}, we explore the impact of the availability of data annotation on our method. As described in Section~\ref{sec:anchor_points}, our method requires references on which to apply the vector field. Specifically, we used 3D bounding boxes to match the objects with the vector fields. While bounding box annotations are often available, this may not always be the case (e.g., SemanticKITTI). To circumvent this, we explore two different strategies: using an off-the-shelf 3D object detector deployed on the training data, or using point-level instance annotations and wrapping each instance in an axis-aligned bounding box.

In the table, we report the effect of both strategies. While the axis-aligned is inherently sub-optimal due to the lack of orientation information, preventing the vector fields from specializing to the object viewing angles, it still brings improvements on the \textit{car} class over the baseline both in-domain and on Waymo. Instead, using the off-the-shelf 3D detector delivers superior performance, especially out-of-domain. In the table, we also explore the impact of the quality of the predictions of the off-the-shelf detector. We do so by considering 10\% less of its bounding box predictions, selected randomly. Remarkably, the performance degrades only slightly for the \textit{car} class, showing the robustness of our method with respect to its assumptions in terms of data annotations.

\setlength{\tabcolsep}{8.3pt}
\begin{sidewaystable}
\begin{center}
\begin{minipage}{\textheight}
\begin{tabular*}{\textheight}{ll|l|cccc|cccc|cccc}
\toprule

\multicolumn{3}{c|}{~} & \multicolumn{4}{c|}{SemanticKITTI} & \multicolumn{4}{c|}{$\rightarrow$ Waymo} 
& \multicolumn{4}{c}{$\rightarrow$ nuScenes}
\\ 
ID & Attack & \multicolumn{1}{l|}{Method} & mIoU & \textit{car} & \textit{pers.} & \textit{bicyc.} & mIoU & \textit{car} & \textit{pers.} & \textit{bicyc.} 
& mIoU & \textit{car} & \textit{pers.} & \textit{bicyc.}
\\
\midrule
c.o & - & Cylinder3D & \textbf{64.39} & 96.02 & 73.84 & 47.15 &   29.43 & 72.10 & 6.69 & 5.68 & 29.97 & 66.04 & 0.00 & \textbf{0.89} \\
u.a & $N\times$a.u.a & [ours] untar.ax.alg. \textit{car}  & 61.82 & \textbf{96.30} & 75.21 & \textbf{50.67} & 28.38 & 74.18 & 12.21 & \textbf{6.49} & 27.75 & 64.88 & 0.04 & 0.58 \\
u.- & $N\times$a.u.- & [ours] untar.-10\% \textit{car}  & 63.92 & 96.16 & 74.02 & 42.04 & 36.65 & 81.25 & 13.15 & 6.17 & 30.42 & 70.01 & \textbf{0.06} & 0.39 \\
u.c & $N\times$a.u.c & [ours] untargeted \textit{car} & 63.85 & 96.03 &	\textbf{77.11} &	33.41 & \textbf{37.74} & \textbf{82.38} & \textbf{17.13} & 5.04 & \textbf{31.55} & \textbf{71.92} & \textbf{0.06} & 0.53 \\
\bottomrule
\end{tabular*}
\caption{Comparison of \protect\culine{3D semantic segmentation} models trained on SemanticKITTI~\cite{behley2019semantickitti} and transferred to Waymo~\cite{sun2020waymo} and nuScenes~\cite{caesar2020nuscenes} validation sets (without fine-tuning). This evaluation assesses the impact of the available data annotation on our method. With u.- we randomly removed 10\% of the bounding boxes predicted by the off-the-shelf 3D detector, so our attack was learned on 10\% less objects, and our model was learned augmenting 10\% less objects. Instead, with u.a we explore the impact of using axis-aligned bounding boxes instead of predicted 3D bounding boxes. The u.c model used throughout this work used all the boxes predicted by the off-the-shelf detector.}
\label{table:annotations}
\end{minipage}
\end{center}
\end{sidewaystable}

\setlength{\tabcolsep}{8pt}
\begin{sidewaystable}
\begin{center}
\begin{minipage}{\textheight}
\begin{tabular*}{\textheight}{ll|cc|cccc|cccc|cccc}
\toprule

& \multicolumn{1}{c|}{~} & \multicolumn{2}{c|}{SemK attack}  & \multicolumn{4}{c|}{SemanticKITTI}  & \multicolumn{4}{c|}{$\rightarrow$ Waymo} 
& \multicolumn{4}{c}{$\rightarrow$ nuScenes}
\\

ID & \multicolumn{1}{l|}{Method} & ~~mIoU & \textit{car} & mIoU & \textit{car} & \textit{pers.} & \textit{bike} & mIoU & \textit{car} & \textit{pers.} & \textit{bike} 
& mIoU & \textit{car} & \textit{pers.} & \textit{bike}
\\

\midrule

c.o & Cylinder3D & - & - & \textbf{64.39} & 96.02 & 73.84 & 47.15 &  29.43 & 72.10 & 6.69 & \textbf{5.68} & 29.97 & 66.04 & 0.00 & \textbf{0.89} \\
u.l & all constr.~no learn & 63.92 & 95.77 &  62.07 & \textbf{96.21} & 75.11 & 48.85 & 28.87 & 78.76 & 12.54 & 4.12 & 27.85 & 63.81 & 0.02 & 0.35 \\
u.u & unleash & \textbf{44.79} & \textbf{27.99} &  61.64 & 96.07 & 72.56 & 46.58 &  30.16 & 69.22 & 13.32 & 5.97  & 30.52 & 63.28 & 0.00 & 0.04 \\
u.r & ray constraint & 52.77 & 61.78 &  62.37 & 95.76 & 71.48 & \textbf{49.26} &  28.61 & 69.35 & 12.46 & 4.30 & 27.41 & 66.59 & \textbf{0.13} & 0.23 \\
u.c & full & 53.25 & 62.61 & 63.85 & 96.03 &	\textbf{77.11} &	33.41 &\textbf{37.74} & \textbf{82.38} & \textbf{17.13} & 5.04 & \textbf{31.55} & \textbf{71.92} & 0.06 & 0.53 \\

\bottomrule

\end{tabular*}
\caption{Ablation on the deformation constraints imposed by our method on \protect\culine{3D semantic segmentation}, compared to Cylinder3D~\cite{zhu2021cylindrical}. All models are based on Cylinder3D, use LiDAR intensity as input, and our adversarial augmentations are untargeted on the \textit{car} class. SemK attack represents the IoUs obtained by Cylinder3D after perturbing the input point clouds with the various configurations.
}
\label{table:ablation_semantic_learning}
\end{minipage}
\end{center}
\end{sidewaystable}

\subsection{Ablation Study on Deformation Constraints}\label{sec:results_ablation_learning}
\textbf{Semantic segmentation} As we introduced sensor-awareness and surface smoothness constraints to our deformations, we investigate their impact in terms of generalization to out-of-domain data. In Table~\ref{table:ablation_semantic_learning}, we report this comparison on 3D semantic segmentation when limiting the deformations to $\epsilon = 30$ cm.
It can be seen that not learning the perturbations (i.e., not adversarial), but applying all our constraints (all constr.~no learn) is already an effective augmentation technique, as it improved on the reference \textit{car} class, also when transferring to Waymo.
Instead, removing all constraints, but learning the vector fields (unleash) led to the strongest adversarial attack (i.e., lowest IoUs after the attack). However, the lack of constraints did not allow for robust generalization, reducing the mIoU in-domain, and the \textit{car} class on Waymo.
When deforming with sensor-awareness (ray constraint), the attack lost effectiveness, but in-domain mIoU increased.
With full, we added the constrain on the surface smoothing (Section~\ref{sec:constraints}), which allowed for superior transfer capabilities, thanks to the improved plausibility of the deformations.


\begin{figure*}[!t]
\centering
  \includegraphics[width=1.0\textwidth]{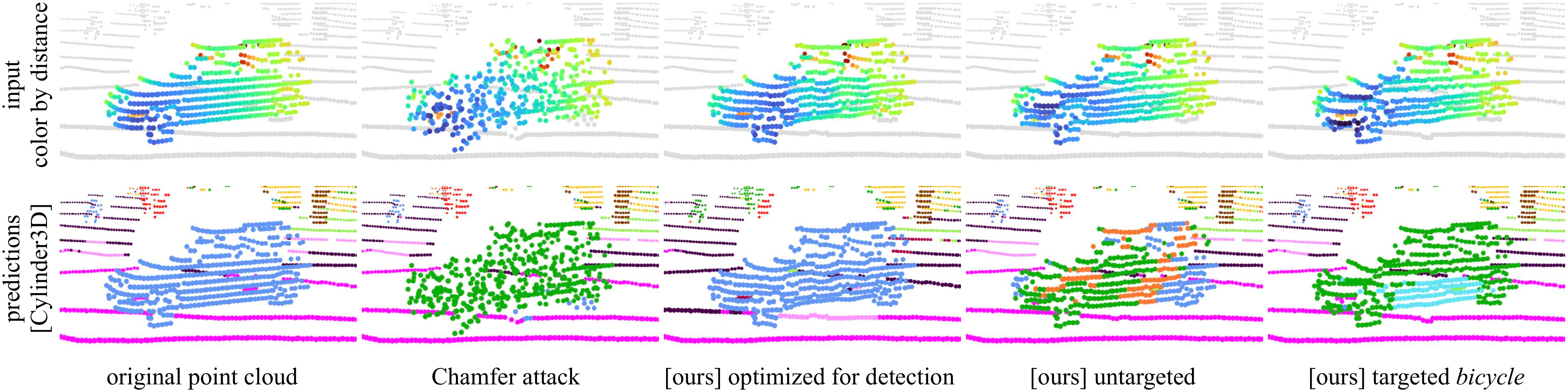}
   \caption{Perturbations and predictions for \protect\uline{3D semantic segmentation}. The first row shows crops of the input point clouds from SemanticKITTI~\cite{behley2019semantickitti}, with a car whose points are colored according to their distance from the LiDAR sensor. Along the row, each point cloud is perturbed by a different adversarial method optimized on \textit{car}, namely the sample-specific Chamfer attack~\cite{liu2020adversarial}, and our approach in three different configurations: optimized for 3D object detection, untargeted, and targeted towards \textit{bicycle} (turquoise in the predictions). The second row shows the predictions of Cylinder3D~\cite{zhu2021cylindrical} on the perturbed point clouds of the first row. All adversarial methods included the intensity signal, except for our vectors optimized for object detection, which did not include it. Therefore, the evaluated model evaluated in the second row takes in input also the intensity values (ID: c.o), except for the one predicting on the point cloud perturbed by the detection vectors (ID: c.n).}
   \label{fig:qualitative_attack_semantic}
\end{figure*}

\section{Qualitative Results}

\subsection{3D Semantic Segmentation, Adversarial Examples}
Figure~\ref{fig:qualitative_attack_semantic} shows the impact of the perturbations applied by our method compared to sample-specific adversarial attacks, represented by the Chamfer attack~\cite{liu2020adversarial}. As discussed in Sections~\ref{sec:results_main_generalization} and~\ref{sec:results_attacks}, by being sample-specific (i.e., instance-specific in our setup), prior works deliver very strong attacks thanks to their highly noticeable perturbations. These substantial perturbations bring their adversarial examples too far from the training distribution (second column of the figure), making the samples unrecognizable. In fact, the attacked car is predicted as \textit{vegetation} (green). However, for the same reasons, augmenting with the point clouds produced by the Chamfer attack (or other sample-specific methods) carries no benefits in terms of generalization, as the data will likely not contain objects of the same category resembling the perturbed objects. Therefore, augmenting with such samples confuses the model.

\subsubsection{Plausibility}
As we focus on improving generalization and robustness to out-of-domain data, our adversarial examples need to be plausible, yet difficult enough for the model to expand the training data distribution. This plausibility can be seen in Figure~\ref{fig:qualitative_attack_semantic} via the significantly less noticeable perturbations applied by our method (from the third to the last columns), compared to the Chamfer attack. Our constraints (e.g., sensor-awareness) mitigated the deformations leading to adversarial examples which still resemble an object of the same class (i.e., \textit{car}). As our plausible adversarial examples are different from the existing training data, they could be similar to real-world out-of-distribution objects (e.g., a damaged car). Therefore, using them as augmentation positively enriches the training data and improves the model generalization and robustness.

\subsubsection{Targeted/Untargeted and Cross-task}
In Figure~\ref{fig:qualitative_attack_semantic}, we compare the effect on the point clouds of three variants of our adversarial method on semantic segmentation, namely the untargeted vectors for the \textit{car} class (a.u.c), the targeted vectors from \textit{car} to \textit{bicycle} (a.t.b), and the vectors optimized for object detection and applied on semantic segmentation. Although the vectors optimized against the 3D detector PointPillars altered the points visually similar to the ones trained for semantic segmentation (targeted and untargeted), their perturbations did not affect Cylinder3D, which managed to accurately segment the deformed car. Conversely, by being optimized for the same task, both our untargeted and targeted vectors perturbed the point cloud in a way that made Cylinder3D wrongly segment the car. On the untargeted adversarial example Cylinder3D predicted a mix of \textit{vegetation} (green), \textit{fence} (orange), and \textit{car} (blue). Instead, on the targeted example Cylinder3D predicted a mix of \textit{bicycle} (turquoise) and \textit{vegetation} (green), as if it recognized a bicycle leaning against a bush, which is rather common in SemanticKITTI~\cite{behley2019semantickitti}. By augmenting with an untargeted sample as the one shown in the figure, the model learns to better distinguish the \textit{car} class from the others. Instead, with a targeted sample, the network improves its decision boundary between the two specific classes, i.e., \textit{car} and \textit{bicycle} as shown. The figure shows both the effectiveness of our adversarial examples, as well as the difference between our targeted and untargeted approach.

\begin{figure*}[t]
\centering
  \includegraphics[width=1.0\textwidth]{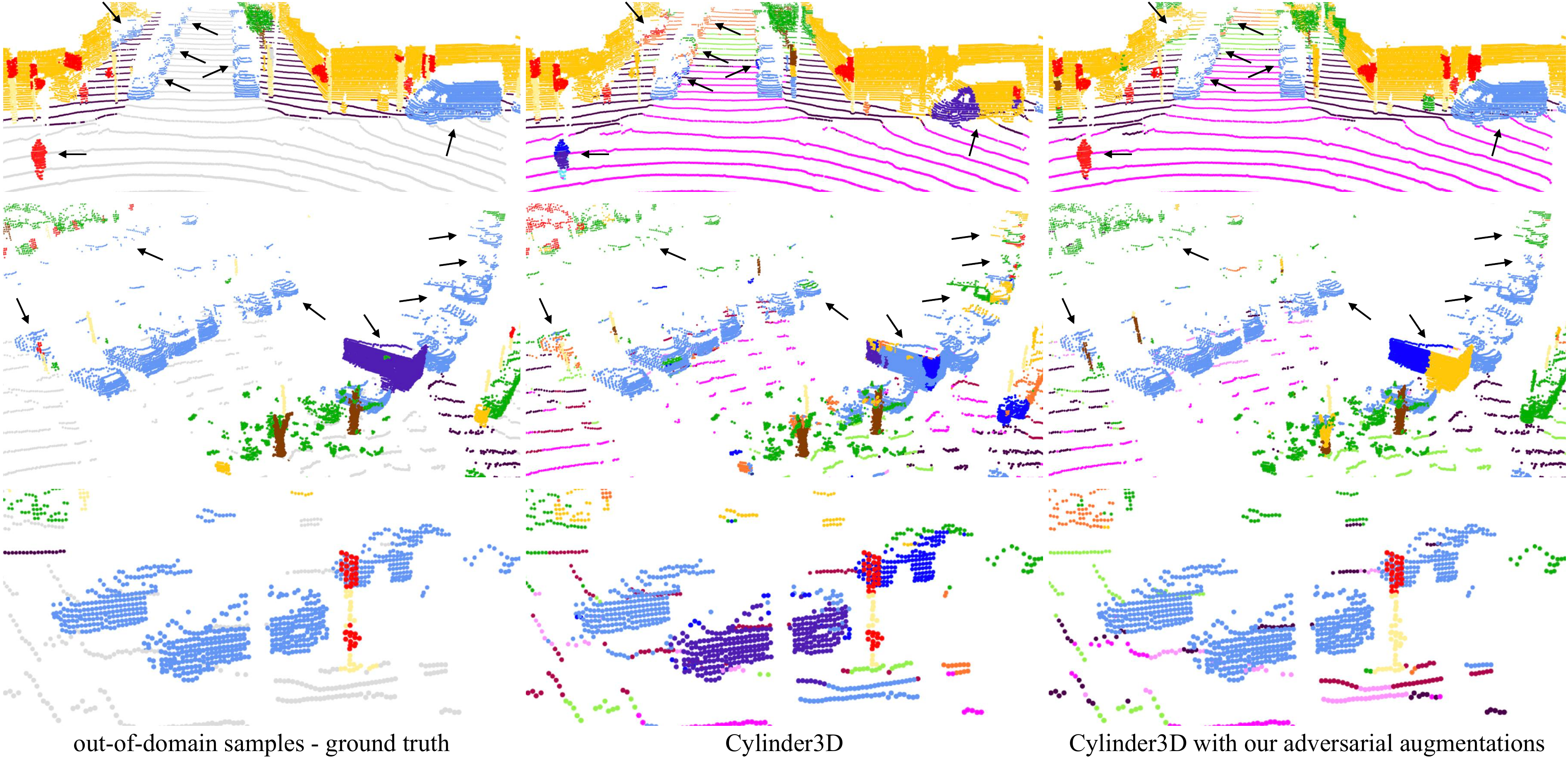}
   \caption{\protect\uline{3D semantic segmentation} predictions on challenging out-of-domain samples from Waymo~\cite{sun2020waymo}. Both models are based on Cylinder3D~\cite{zhu2021cylindrical}, trained only on SemanticKITTI, and transferred to Waymo without fine-tuning. Two Cylinder3D models are compared: without (c.o) and with (u.c) our adversarial augmentations. Black arrows highlight some of the objects segmented wrongly by the baseline. The road and other classes were ignored in the transfer to Waymo due to misaligned definitions across the datasets.}
   \label{fig:qualitative_semantic}
\end{figure*}

\subsubsection{Limitations}
As we aimed at preserving the comparability between our detection and semantic vectors for cross-task comparisons, we did not selectively perturb only the points in the 3D bounding box corresponding to the class of interest (i.e., \textit{car}). Instead, we perturb all the points inside the box. This is visible in Figure~\ref{fig:qualitative_attack_semantic} on a portion of the road next to the car, which was shifted by the vectors. Furthermore, by relying on predicted bounding boxes for cars, we could not perturb points outside the boxes, despite being part of the same instance. In the example shown in Figure~\ref{fig:qualitative_attack_semantic}, the off-the-shelf detector used (\cite{deng2021voxelrcnn}) did not fully detect the car, and left a few points outside the box. This is visible on the roof line and along other lines of points on the side of the car. Therefore, these points were left unperturbed by all methods, including the Chamfer attack (which was applied on the same objects). Nevertheless, in the adversarial examples those points were anyways wrongly segmented by Cylinder3D.

\begin{figure*}[t]
\centering
  \includegraphics[width=1.0\textwidth]{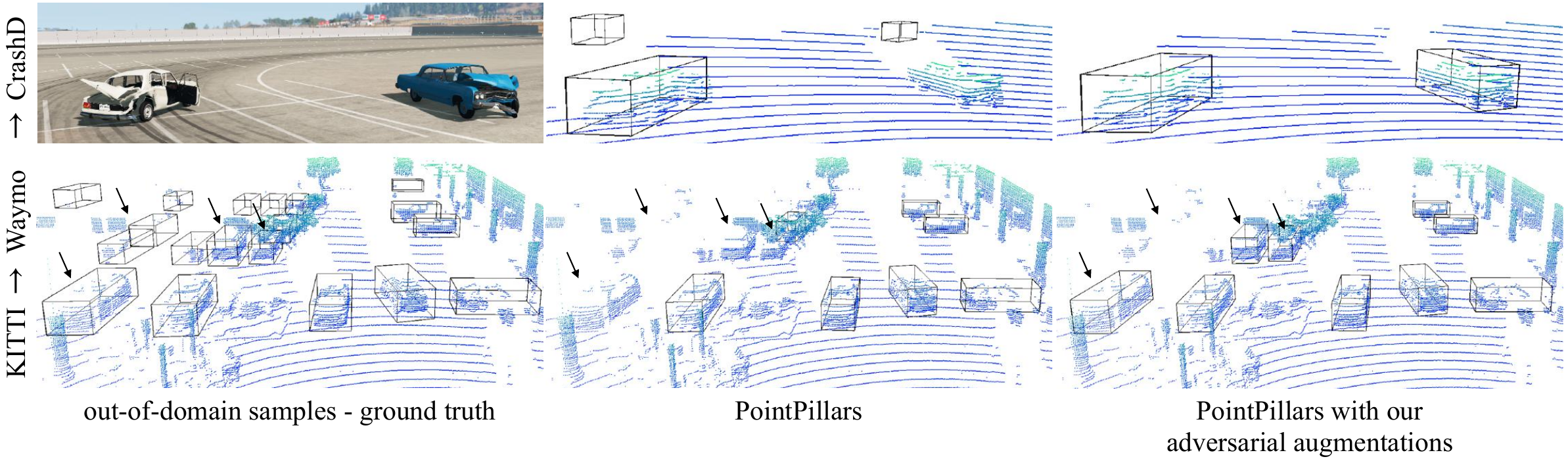}
   \caption{
   \protect\uline{3D object detection} predictions on challenging out-of-domain samples from the proposed CrashD dataset (top) and Waymo~\cite{sun2020waymo} (bottom). Both models are based on PointPillars~\cite{lang_pointpillars_2019}, trained only on KITTI, and transferred to CrashD and Waymo without fine-tuning. Two PointPillars models are compared: without (p.p) and with (p.o) our adversarial augmentations. Black arrows highlight some of the objects segmented wrongly by the baseline.}
   \label{fig:qualitative_detection}
\end{figure*}

\subsection{3D Semantic Segmentation, Robustness and Generalization}
Figure~\ref{fig:qualitative_semantic} shows the semantic predictions of Cylinder3D trained with and without our adversarial augmentations. Both models are trained on SemanticKITTI, and transferred to the point clouds of Waymo shown in the figure. As the adversarial examples used to train our model were only on cars, we focus on the \textit{car} class. Due to the large domain gap between the two datasets, neither of the models was able to properly segment all cars in the scenes. However, while ours missed only the cars with few points, the standard Cylinder3D could not segment several visible ones, confusing them with \textit{truck}, \textit{building}, \textit{vegetation}, and other classes. The superior performance of our model across the various scenes confirms the added robustness and generalization of our method towards challenging out-of-domain data.

\subsection{3D Object Detection}
In Figure~\ref{fig:qualitative_detection} we compare the transfer from KITTI to CrashD and Waymo~\cite{sun2020waymo} in terms of 3D bounding boxes predicted by PointPillars~\cite{lang_pointpillars_2019} trained with and without our adversarial augmentations. For the proposed CrashD, our approach correctly identified both damaged cars, without false positives. The figure testifies also the intensity of the \textit{hard} damages of CrashD, and how challenging these vehicles are for a 3D detector compared to standard cars (e.g., KITTI), resembling natural adversarial examples. For the transfer from KITTI to Waymo, both the baseline and our augmented model had difficulty detecting all cars in the scene, especially those with fewer points in the parking lot on the left of the crops. However, the standard PointPillars ignored also 3 recognizable cars featuring lots of points. Instead, albeit leaving room for improvement, ours could recognize them.

We refer to the \textbf{Supplementary Material} for more results on indoor settings, the IoUs on the complete set of classes, as well as details on the class mappings for the transfers to Waymo and nuScenes.


\section{Conclusion}
In this work we presented a flexible adversarial augmentation approach, which improves generalization and robustness across multiple 3D tasks. By expanding the available training data with plausible adversarial examples, our augmentations increase safety by addressing challenging long tail and out-of-domain samples without the burden of capturing them in the real-world. As extensive experiments across multiple tasks and datasets showed its benefits and flexibility, we believe our method constitutes a valuable step towards safe and robust perception for high automation systems.

\textbf{Acknowledgments}
This work was partly sponsored by the German Federal Ministry for Economic Affairs and Energy (grant number 19A19005B), through the VDA KI-Absicherung project.



\appendix

\section{Appendix}
In this supplementary material we include further details and results. Specifically, implementation details on compared adversarial methods, specifications of the dataset transfers,
and additional quantitative results including all evaluated classes for completeness.

\subsection{Additional Details}

\subsubsection{Iterative gradient L2 attack}
As in our conference publication~\cite{lehner20223dvfield}, for the iterative gradient L2 attack~\cite{xiang_generating_2019} we minimized our adversarial loss $\mathcal{L}_{\rm adv}$ constraining the deformation $\bm{m}$ for each point $\bm{p}$ with $\Vert \bm{m} \Vert_{2} < \epsilon$, with $\epsilon = 30$ cm. The same holds for the resulting intensity change $\Lambda$ with $\Vert \Lambda \Vert_{2} < \psi$, with $\psi = 0.3$.

\setlength{\tabcolsep}{4.45pt}
\begin{sidewaystable}
\begin{center}
\begin{minipage}{\textheight}
\begin{tabular*}{\textheight}{l|cccccccccccccccccccc}
\toprule

\multicolumn{1}{c|}{~} & \multicolumn{20}{c}{SemanticKITTI} \\ 
\\

ID & \begin{sideways}mIoU\end{sideways} & \begin{sideways}\textit{car}\end{sideways} & \begin{sideways}\textit{truck}\end{sideways} & \begin{sideways}\textit{bicycle}\end{sideways} & \begin{sideways}\textit{motorcycle}\end{sideways} & \begin{sideways}\textit{other vehicle}\end{sideways} & \begin{sideways}\textit{person}\end{sideways} & \begin{sideways}\textit{bicyclist}\end{sideways} & \begin{sideways}\textit{motorcyclist}\end{sideways} & \begin{sideways}\textit{road}\end{sideways} & \begin{sideways}\textit{sidewalk}\end{sideways} & \begin{sideways}\textit{parking}\end{sideways} & \begin{sideways}\textit{other ground}\end{sideways} & \begin{sideways}\textit{building}\end{sideways} & \begin{sideways}\textit{vegetation}\end{sideways} & \begin{sideways}\textit{trunk}\end{sideways} & \begin{sideways}\textit{terrain}\end{sideways} & \begin{sideways}\textit{fence}\end{sideways} & \begin{sideways}\textit{pole}\end{sideways} & \begin{sideways}\textit{traffic sign}\end{sideways} \\

\midrule

c.n & 59.23 & \textbf{96.23} & \textbf{32.61} & 37.42 & \textbf{83.07} & \textbf{57.53} & 68.67 & 80.01 & 0.00 & 92.73 & 34.88 & 77.91 & 0.58 & 87.48 & 45.10 & \textbf{86.58} & \textbf{64.71} & \textbf{71.64} & \textbf{61.84} & \textbf{46.42} \\

u.n & \textbf{59.40} & 95.73 & 26.53 & \textbf{43.69} & 79.96 & 53.79 & \textbf{70.35} & \textbf{85.00} & 0.00 & \textbf{92.84} & \textbf{38.54} & \textbf{78.16} & \textbf{5.35} & \textbf{89.18} & \textbf{53.62} & 85.07 & 59.52 & 66.39 & 60.58 & 44.31 \\

\cmidrule(lr){1-21}

c.o & \textbf{64.39} & 96.02 & \textbf{47.15} & \textbf{69.11} & 88.96 & 54.52 & 73.84 & 84.72 & 0.05 & 94.40 & \textbf{47.40} & 80.54 & \textbf{1.66} & 89.38 & 54.19 & \textbf{87.44} & \textbf{67.18} & \textbf{72.83} & 63.95 & \textbf{50.02} \\

u.c & 63.85 & \textbf{96.03} & 33.41 & 68.24 & \textbf{89.03} & \textbf{59.32} & \textbf{77.11} & \textbf{91.82} & \textbf{0.06} & \textbf{94.41} & 40.08 & \textbf{81.00} & 0.30 & \textbf{90.27} & \textbf{57.43} & 85.33 & 66.87 & 67.57 & \textbf{65.20} & 49.64 \\

\bottomrule

\end{tabular*}
\caption{Results on the SemanticKITTI validation set~\cite{behley2019semantickitti} of \protect\culine{3D semantic segmentation} models. All classes are reported. IDs are consistent with the main manuscript.}
\label{table:all_classes_SK}
\end{minipage}
\end{center}
\end{sidewaystable}

\begin{table*}[t]
\begin{center}
\begin{tabular}{l|ccccccccccccc}
\toprule

\multicolumn{1}{c|}{~} & \multicolumn{13}{c}{$\rightarrow Waymo$} \\ 
\\

ID & \begin{sideways}mIoU\end{sideways} & \begin{sideways}\textit{car}\end{sideways} & \begin{sideways}\textit{bicycle}\end{sideways} & \begin{sideways}\textit{motorcycle}\end{sideways} & \begin{sideways}\textit{truck}\end{sideways} & \begin{sideways}\textit{other-vehicle}\end{sideways} & \begin{sideways}\textit{person}\end{sideways} & \begin{sideways}\textit{sidewalk}\end{sideways} & \begin{sideways}\textit{building}\end{sideways} & \begin{sideways}\textit{vegetation}\end{sideways} & \begin{sideways}\textit{trunk}\end{sideways} & \begin{sideways}\textit{pole}\end{sideways} & \begin{sideways}\textit{traffic-sign}\end{sideways}  \\

\midrule

c.n & 37.39 & 63.74 & 4.05 & 4.93 & \textbf{17.29} & 8.86 & 37.95 & \textbf{68.74} & 77.74 & 78.18 & 38.66 & \textbf{19.55} & 28.97 \\

u.n & \textbf{40.36} & \textbf{70.70} & \textbf{7.57} & \textbf{8.29} & 12.12 & \textbf{13.35} & \textbf{52.73} & 63.93 & \textbf{79.41} & \textbf{80.09} & \textbf{40.98} & 19.13 & \textbf{36.05} \\

\cmidrule(lr){1-14}

c.o & 29.43 & 72.10 & \textbf{5.68} & 1.92 & \textbf{12.02} & 7.99 & 6.69 & 42.47 & 66.00 & 51.51 & 12.45 & \textbf{38.12} & 36.15 \\

u.c & \textbf{37.74} & \textbf{82.38} & 5.04 & \textbf{2.70} & \textbf{9.23} & \textbf{20.74} & \textbf{17.13} & \textbf{74.28} & \textbf{73.32} & \textbf{70.16} & \textbf{15.72} & 36.47 & \textbf{45.77} \\

\bottomrule

\end{tabular}
\caption{Transfer results on the Waymo~\cite{sun2020waymo} validation set of \protect\culine{3D semantic segmentation} models trained on SemanticKITTI~\cite{behley2019semantickitti}. All classes are reported. IDs are consistent with the main manuscript.}
\label{table:all_classes_W}
\end{center}
\end{table*}

\begin{table*}[t]
\begin{center}
\begin{tabular}{l|ccccccccc}
\toprule

\multicolumn{1}{c|}{~} & \multicolumn{9}{c}{$\rightarrow nuScenes$} \\ 
\\

ID & \begin{sideways}mIoU\end{sideways} & \begin{sideways}\textit{car}\end{sideways} & \begin{sideways}\textit{bicycle}\end{sideways} & \begin{sideways}\textit{truck}\end{sideways} & \begin{sideways}\textit{person}\end{sideways} & \begin{sideways}\textit{road}\end{sideways} & \begin{sideways}\textit{sidewalk}\end{sideways} & \begin{sideways}\textit{vegetation}\end{sideways} & \begin{sideways}\textit{terrain}\end{sideways}  \\

\midrule

c.n & 32.11 & 56.17 & 0.35 & 0.23 & 0.04 & \textbf{78.81} & \textbf{33.89} & \textbf{71.98} & 15.44 \\
u.n & \textbf{34.30} & \textbf{63.13} & \textbf{2.52} & \textbf{2.59} & \textbf{1.02} & 77.15 & 33.65 & 68.64 & \textbf{25.70} \\

\cmidrule(lr){1-10}

c.o & 29.97 & 66.04 & \textbf{0.89} & 0.05 & 0.00 & 75.79 & 28.00 & 57.94 & 11.08 \\

u.c & \textbf{31.55} & \textbf{71.92} & 0.53 & \textbf{0.12} & \textbf{0.06} & \textbf{77.27} & \textbf{30.19} & \textbf{59.61} & \textbf{12.70} \\

\bottomrule

\end{tabular}
\caption{Transfer results on the nuScenes validation set~\cite{caesar2020nuscenes} of \protect\culine{3D semantic segmentation} models trained on SemanticKITTI~\cite{behley2019semantickitti}. All classes are reported. IDs are consistent with the main manuscript.}
\label{table:all_classes_nS}
\end{center}
\end{table*}


\subsubsection{Chamfer attack}
The Chamfer attack~\cite{liu2020adversarial} is based on the Chamfer distance, which is measured as the average of the sum of the distances between each point in the original point cloud and its nearest neighbor in the deformed one. Therefore, using this distance function encourages points to move along the object surface.
As in our conference publication~\cite{lehner20223dvfield}, we used the Chamfer distance to measure the gap between the original and perturbed point clouds, which is given by:
\begin{equation}
\mathcal{C}(X, Y) = \frac{1}{|X|}\sum_{x \in X} \min_{y \in Y} ||x - y||_2
\end{equation}
for two sets $X$ and $Y$.
We perturbed by minimizing:
\begin{equation}\label{chamfer}
\mathcal{L}_{\rm cha} = \mathcal{L}_{\rm adv} + \lambda \mathcal{C}(\bm{p} + \bm{m}, \bm{p})
\end{equation}
with $\lambda$ set to 0.1 and the amount of deformation constrained by $\mathcal{C}(\bm{p} + \bm{m}, \bm{p}) < \epsilon$, with $\epsilon = 30$ cm. If we also attack the intensity, we add the term $\lambda \mathcal{C}(\bm{p} + \Lambda, \bm{p})$ to eqation~\ref{chamfer} which is also constrained by $\mathcal{C}(\bm{p} + \Lambda, \bm{p})< \psi$, with $\psi = 0.3$. It should be noted that single deformations vectors could lead to perturbations larger than 30 cm, since what is bounded is the overall Chamfer distance and not single vectors. This attack led to only a small amount of perturbed points, but the ones that moved showed large displacements.

\subsubsection{Adversarial removal}
For the removal attack we followed \cite{yang2019adversarial_removal} and removed 10\% of the \textit{critical points} of an object, that have the greatest effect on the prediction of an object when changed. As we did for our conference publication~\cite{lehner20223dvfield}, we estimated them as those with the highest deformation magnitude from the iterative gradient L2 attack~\cite{xiang_generating_2019}.

\subsubsection{Adversarial generation}
As in our conference publication~\cite{lehner20223dvfield}, we followed \cite{xiang_generating_2019} adding 10\% of the objects points and  initialized them at the location of the \textit{critical points} (see removal). We then performed the iterative gradient L2 attack~\cite{xiang_generating_2019} only on the added points, thus shifting their 3D position and also intensity (if applicable) to decrease the models prediction.

\subsubsection{Dataset Transfers}
In this section we describe the class mappings used for transferring from SemanticKITTI to Waymo or nuScenes.
When transferring to Waymo we considered the following 12 classes: \textit{car}, \textit{truck}, \textit{other-vehicle} merged with \textit{bus} (as SemanticKITTI does not distinguish them), \textit{pedestrian} (\textit{person} in SemanticKITTI), \textit{sign} (\textit{traffic-sign} in SemanticKITTI), \textit{pole}, \textit{bicycle}, \textit{motorcycle}, \textit{building}, \textit{vegetation}, \textit{trunk}, and \textit{sidewalk}. We excluded the remaining classes due to incompatible definitions across the datasets, e.g., \textit{road} in Waymo includes driveways going over sidewalks while this is not the case in SemanticKITTI.
Instead, for transfers to nuScenes we considered the following 8 classes: \textit{human.pedestrian} \textit{adult} combined with \textit{child}, \textit{construction\_worker}, \textit{police\_officer} (overall considered as \textit{person} in SemanticKITTI), then \textit{bicycle} (includes also \textit{bicyclist} from SemanticKITTI), \textit{car}, \textit{truck}, \textit{driveable\_surface} (includes \textit{road} and \textit{parking} from SemanticKITTI), \textit{sidewalk}, \textit{terrain}, and \textit{vegetation} (includes also \textit{trunk} from SemanticKITTI). We ignored the remaining classes due to contrasting labeling definitions. In nuScenes, \textit{motorcycle} includes 3-wheeled vehicles such as auto rickshaws, which would be considered \textit{other-vehicle} in SemanticKITTI. Moreover, in SemanticKITTI police cars would count as \textit{car}, while they have a dedicated class in nuScenes, i.e., \textit{vehicle.emergency.police}.
Furthermore, when comparing with prior works, we followed their class mapping~\cite{yi2021complete_and_label}. Towards this end, for the 2-class setup on Waymo we followed~\cite{yi2021complete_and_label} and considered only \textit{person} and the super-class \textit{vehicle}. The latter is the combination of \textit{car}, \textit{bicycle}, \textit{motorcycle}, \textit{truck}, \textit{other-vehicle}, and \textit{bus}. For nuScenes, for the 10-class we followed~\cite{yi2021complete_and_label}, who added the following on top of our mapping: \textit{motorcycle} and \textit{other-vehicle}. The latter is the combination of: \textit{vehicle.bus} (\textit{bendy} and \textit{rigid}), \textit{vehicle.construction}, \textit{vehicle.emergency} (\textit{ambulance} and \textit{police}), and \textit{vehicle.trailer}. It should be noted that this 10-class mapping by~\cite{yi2021complete_and_label} introduces inconsistencies, e.g., \textit{vehicle.trailer} in nuScenes includes also trailers attached to trucks, however, these are predicted as \textit{truck} in nuScenes, because in SemanticKITTI they are considered part of \textit{truck}. Nevertheless, we followed their mapping, but only when comparing to their work.
Lastly, we did not alter our models or their training procedures towards the transfers. All our segmentation models are trained on the full 19 classes of SemanticKITTI. The mappings above are used only at evaluation time to match the annotations of the datasets. This is different from the approach of~\cite{yi2021complete_and_label}, who trained on the 10 classes.

\subsection{Additional Results}
For completeness, in Tables~\ref{table:all_classes_SK},~\ref{table:all_classes_W}, and~\ref{table:all_classes_nS} we report the IoU for all classes used in our main 3D semantic segmentation experiments for SemanticKITTI, Waymo, and nuScenes respectively. These tables complement the ones in the main manuscript, where we selected a few representative classes.

Furthermore, in our conference publication and its Supplementary Material~\cite{lehner20223dvfield}, we report additional experiments, including extensive results on the proposed CrashD dataset, various ablation studies on the proposed attack and augmentation methods, as well as quantitative and qualitative results on the indoor SUN RGB-D dataset.

\bibliographystyle{spbasic}      
\bibliography{bib}   

\end{document}